\documentclass[english,ruled]{article}
\usepackage[T1]{fontenc}
\usepackage[latin9]{inputenc}
\usepackage{verbatim}
\usepackage[algo2e,ruled]{algorithm2e}
\usepackage{algorithm}
\usepackage{amsmath}
\usepackage{amsthm}
\usepackage{amssymb}
\usepackage{graphicx}
\usepackage{geometry}
\usepackage{enumitem}
\makeatletter
\usepackage[toc,page,header]{appendix}
\usepackage{minitoc}
\usepackage{comment}
\usepackage{ifthen}
\newboolean{doublecolumn}
\setboolean{doublecolumn}{true}
\usepackage{multirow}
\usepackage[colorlinks]{hyperref}
\usepackage{booktabs}       
\usepackage{amsfonts}       
\usepackage{nicefrac}       
\usepackage{authblk}
\usepackage{optidef}
\usepackage{array}
\usepackage{babel}
\usepackage[resetlabels]{multibib}
\usepackage{cite}
\usepackage{subcaption}
\usepackage{ragged2e}
\usepackage{titletoc}
\usepackage{bm}
\usepackage{makecell}
\usepackage{geometry}                		
\usepackage{graphicx}				
\usepackage{amssymb}
\usepackage[dvipsnames]{xcolor}
\usepackage[capitalise]{cleveref}

\newcolumntype{L}[1]{>{\raggedright\let\newline\\\arraybackslash\hspace{0pt}}m{#1}}
\newcolumntype{C}[1]{>{\centering\let\newline\\\arraybackslash\hspace{0pt}}m{#1}}
\newcolumntype{R}[1]{>{\raggedleft\let\newline\\\arraybackslash\hspace{0pt}}m{#1}}

\SetKw{Return}{\textbf{Return:}}

\renewcommand \partname{}

\theoremstyle{plain}
\newtheorem{lem}{\protect\lemmaname}[section]
\theoremstyle{remark}
\newtheorem{rem}{\protect\remarkname}[section]
\theoremstyle{plain}
\newtheorem{thm}{\protect\theoremname}[section]
\theoremstyle{plain}

\providecommand{\corollaryname}{Corollary}
\theoremstyle{plain}
\newtheorem{coro}{\protect\corollaryname}[section]
\newtheorem{ass}{Assumption}

\providecommand{\propositionname}{Proposition}
\providecommand{\lemmaname}{Lemma}
\providecommand{\remarkname}{Remark}
\providecommand{\theoremname}{Theorem}

\newcommand{\E}{\mathbb E}
\newcommand{\deff}{d_{\textup{eff}}}
\newcommand{\bigO}{\mathcal O}
\newcommand{\bigOt}{\tilde {\mathcal O}}
\newcommand{\lmax}{\lambda_{\textup{max}}}

\newcommand{\smax}{\sigma_{\textup{max}}}
\newcommand{\smin}{\sigma_{\textup{min}}}
\newcommand{\Nstar}{\mathcal{N}_{\varepsilon_0}(w_\star)}
\newcommand{\Lc}{\mathcal L}
\newcommand{\Rc}{\mathcal R}
\newcommand{\q}{\mathfrak{q}}
\newcommand{\qmax}{\mathfrak{q}_{\textup{max}}}

\newcommand{\sapph}{\texttt{SAPPHIRE}}

\newcommand{\condsmall}{\ifthenelse{\boolean{doublecolumn}}{\small}{}}

\makeatother

\newcites{app}{References in the Appendix}

\author{Jingruo Sun\thanks{\url{jingruo@stanford.edu}}~}
\author{Zachary Frangella\thanks{\url{zfran@stanford.edu}}~}
\author{Madeleine Udell\thanks{\url{udell@stanford.edu}}~}

\affil{Management Science and Engineering, Stanford University}

\global\long\def\assign{\coloneqq}%

\newtheorem{definition}{Definition}

\title{\textbf{SAPPHIRE: Preconditioned Stochastic Variance Reduction for Faster Large-Scale Statistical Learning}}
\date{}

\begin{document}

\global\long\def\inprod#1#2{\left\langle #1,#2\right\rangle }%

\global\long\def\inner#1#2{\langle#1,#2\rangle}%

\global\long\def\binner#1#2{\big\langle#1,#2\big\rangle}%

\global\long\def\Binner#1#2{\Big\langle#1,#2\Big\rangle}%

\global\long\def\norm#1{\lVert#1\rVert}%

\global\long\def\bnorm#1{\big\Vert#1\big\Vert}%

\global\long\def\Bnorm#1{\Big\Vert#1\Big\Vert}%

\global\long\def\setnorm#1{\Vert#1\Vert_{-}}%

\global\long\def\bsetnorm#1{\big\Vert#1\big\Vert_{-}}%

\global\long\def\Bsetnorm#1{\Big\Vert#1\Big\Vert_{-}}%


\global\long\def\brbra#1{\big(#1\big)}%
\global\long\def\Brbra#1{\Big(#1\Big)}%
\global\long\def\rbra#1{(#1)}%


\global\long\def\sbra#1{[#1]}%
\global\long\def\bsbra#1{\big[#1\big]}%
\global\long\def\Bsbra#1{\Big[#1\Big]}%


\global\long\def\cbra#1{\{#1\}}%
\global\long\def\bcbra#1{\big\{#1\big\}}%
\global\long\def\Bcbra#1{\Big\{#1\Big\}}%
\global\long\def\vertiii#1{\left\vert \kern-0.25ex  \left\vert \kern-0.25ex  \left\vert #1\right\vert \kern-0.25ex  \right\vert \kern-0.25ex  \right\vert }%

\global\long\def\matr#1{\bm{#1}}%

\global\long\def\til#1{\tilde{#1}}%

\global\long\def\wtil#1{\widetilde{#1}}%

\global\long\def\wh#1{\widehat{#1}}%

\global\long\def\mcal#1{\mathcal{#1}}%

\global\long\def\mbb#1{\mathbb{#1}}%

\global\long\def\mtt#1{\mathtt{#1}}%

\global\long\def\ttt#1{\texttt{#1}}%

\global\long\def\dtxt{\textrm{d}}%

\global\long\def\bignorm#1{\bigl\Vert#1\bigr\Vert}%

\global\long\def\Bignorm#1{\Bigl\Vert#1\Bigr\Vert}%

\global\long\def\rmn#1#2{\mathbb{R}^{#1\times#2}}%

\global\long\def\deri#1#2{\frac{d#1}{d#2}}%

\global\long\def\pderi#1#2{\frac{\partial#1}{\partial#2}}%

\global\long\def\limk{\lim_{k\rightarrow\infty}}%

\global\long\def\trans{\textrm{T}}%

\global\long\def\onebf{\mathbf{1}}%

\global\long\def\zerobf{\mathbf{0}}%

\global\long\def\zero{\bm{0}}%


\global\long\def\Euc{\mathrm{E}}%

\global\long\def\Expe{\mathbb{E}}%

\global\long\def\rank{\mathrm{rank}}%

\global\long\def\range{\mathrm{range}}%

\global\long\def\diam{\mathrm{diam}}%

\global\long\def\epi{\mathrm{epi} }%

\global\long\def\relint{\mathrm{relint} }%

\global\long\def\inte{\operatornamewithlimits{int}}%

\global\long\def\cov{\mathrm{Cov}}%

\global\long\def\argmin{\operatornamewithlimits{arg\,min}}%

\global\long\def\argmax{\operatornamewithlimits{arg\,max}}%

\global\long\def\tr{\operatornamewithlimits{tr}}%

\global\long\def\dis{\operatornamewithlimits{dist}}%

\global\long\def\sign{\operatornamewithlimits{sign}}%

\global\long\def\prob{\mathrm{Prob}}%

\global\long\def\spans{\textrm{span}}%

\global\long\def\st{\operatornamewithlimits{s.t.}}%

\global\long\def\dom{\mathrm{dom}}%

\global\long\def\prox{\mathrm{prox}}%

\global\long\def\for{\mathrm{for}}%

\global\long\def\diag{\mathrm{diag}}%

\global\long\def\and{\mathrm{and}}%

\global\long\def\st{\mathrm{s.t.}}%

\global\long\def\dist{\mathrm{dist}}%

\global\long\def\Var{\operatornamewithlimits{Var}}%

\global\long\def\raw{\rightarrow}%

\global\long\def\law{\leftarrow}%

\global\long\def\Raw{\Rightarrow}%

\global\long\def\Law{\Leftarrow}%

\global\long\def\vep{\varepsilon}%

\global\long\def\dom{\operatornamewithlimits{dom}}%

\global\long\def\tsum{{\textstyle {\sum}}}%

\global\long\def\Cbb{\mathbb{C}}%

\global\long\def\Ebb{\mathbb{E}}%

\global\long\def\Fbb{\mathbb{F}}%

\global\long\def\Nbb{\mathbb{N}}%

\global\long\def\Rbb{\mathbb{R}}%

\global\long\def\extR{\widebar{\mathbb{R}}}%

\global\long\def\Pbb{\mathbb{P}}%

\global\long\def\Mrm{\mathrm{M}}%

\global\long\def\Acal{\mathcal{A}}%

\global\long\def\Bcal{\mathcal{B}}%

\global\long\def\Ccal{\mathcal{C}}%

\global\long\def\Dcal{\mathcal{D}}%

\global\long\def\Ecal{\mathcal{E}}%

\global\long\def\Fcal{\mathcal{F}}%

\global\long\def\Gcal{\mathcal{G}}%

\global\long\def\Hcal{\mathcal{H}}%

\global\long\def\Ical{\mathcal{I}}%

\global\long\def\Kcal{\mathcal{K}}%

\global\long\def\Lcal{\mathcal{L}}%

\global\long\def\Mcal{\mathcal{M}}%

\global\long\def\Ncal{\mathcal{N}}%

\global\long\def\Ocal{\mathcal{O}}%

\global\long\def\Pcal{\mathcal{P}}%

\global\long\def\Scal{\mathcal{S}}%

\global\long\def\Tcal{\mathcal{T}}%

\global\long\def\Wcal{\mathcal{W}}%

\global\long\def\Xcal{\mathcal{X}}%

\global\long\def\Ycal{\mathcal{Y}}%

\global\long\def\Zcal{\mathcal{Z}}%

\global\long\def\i{i}%


\global\long\def\abf{\mathbf{a}}%

\global\long\def\bbf{\mathbf{b}}%

\global\long\def\cbf{\mathbf{c}}%

\global\long\def\dbf{\mathbf{d}}%

\global\long\def\fbf{\mathbf{f}}%

\global\long\def\lambf{\bm{\lambda}}%

\global\long\def\alphabf{\bm{\alpha}}%

\global\long\def\sigmabf{\bm{\sigma}}%

\global\long\def\thetabf{\bm{\theta}}%

\global\long\def\deltabf{\bm{\delta}}%

\global\long\def\sbf{\mathbf{s}}%

\global\long\def\lbf{\mathbf{l}}%

\global\long\def\ubf{\mathbf{u}}%

\global\long\def\vbf{\mathbf{v}}%

\global\long\def\wbf{\mathbf{w}}%

\global\long\def\xbf{\mathbf{x}}%

\global\long\def\ybf{\mathbf{y}}%

\global\long\def\zbf{\mathbf{z}}%

\global\long\def\Abf{\mathbf{A}}%

\global\long\def\Ubf{\mathbf{U}}%

\global\long\def\Pbf{\mathbf{P}}%

\global\long\def\Ibf{\mathbf{I}}%

\global\long\def\Ebf{\mathbf{E}}%

\global\long\def\Mbf{\mathbf{M}}%

\global\long\def\Qbf{\mathbf{Q}}%

\global\long\def\Lbf{\mathbf{L}}%

\global\long\def\Pbf{\mathbf{P}}%


\global\long\def\abm{\bm{a}}%

\global\long\def\bbm{\bm{b}}%

\global\long\def\cbm{\bm{c}}%

\global\long\def\dbm{\bm{d}}%

\global\long\def\ebm{\bm{e}}%

\global\long\def\fbm{\bm{f}}%

\global\long\def\gbm{\bm{g}}%

\global\long\def\hbm{\bm{h}}%

\global\long\def\pbm{\bm{p}}%

\global\long\def\qbm{\bm{q}}%

\global\long\def\rbm{\bm{r}}%

\global\long\def\sbm{\bm{s}}%

\global\long\def\tbm{\bm{t}}%

\global\long\def\ubm{\bm{u}}%

\global\long\def\vbm{\bm{v}}%

\global\long\def\wbm{\bm{w}}%

\global\long\def\xbm{\bm{x}}%

\global\long\def\ybm{\bm{y}}%

\global\long\def\zbm{\bm{z}}%

\global\long\def\Abm{\bm{A}}%

\global\long\def\Bbm{\bm{B}}%

\global\long\def\Cbm{\bm{C}}%

\global\long\def\Dbm{\bm{D}}%

\global\long\def\Ebm{\bm{E}}%

\global\long\def\Fbm{\bm{F}}%

\global\long\def\Gbm{\bm{G}}%

\global\long\def\Hbm{\bm{H}}%

\global\long\def\Ibm{\bm{I}}%

\global\long\def\Jbm{\bm{J}}%

\global\long\def\Lbm{\bm{L}}%

\global\long\def\Obm{\bm{O}}%

\global\long\def\Pbm{\bm{P}}%

\global\long\def\Qbm{\bm{Q}}%

\global\long\def\Rbm{\bm{R}}%

\global\long\def\Ubm{\bm{U}}%

\global\long\def\Vbm{\bm{V}}%

\global\long\def\Wbm{\bm{W}}%

\global\long\def\Xbm{\bm{X}}%

\global\long\def\Ybm{\bm{Y}}%

\global\long\def\Zbm{\bm{Z}}%

\global\long\def\lambm{\bm{\lambda}}%

\global\long\def\alphabm{\bm{\alpha}}%

\global\long\def\albm{\bm{\alpha}}%

\global\long\def\taubm{\bm{\tau}}%

\global\long\def\mubm{\bm{\mu}}%

\global\long\def\yrm{\mathrm{y}}%

\newcommand{\R}{\mathbb{R}}
\maketitle

\begin{abstract} \label{1. abstract}

Regularized empirical risk minimization (rERM) has become important in data-intensive fields such as genomics and advertising,
with stochastic gradient methods typically used to solve the largest problems.
However, ill-conditioned objectives and non-smooth regularizers undermine the performance of traditional stochastic gradient methods, leading to slow convergence and significant computational costs. 
To address these challenges, we propose the \sapph{} (\textbf{S}ketching-based \textbf{A}pproximations for \textbf{P}roximal \textbf{P}reconditioning and \textbf{H}essian \textbf{I}nexactness with Variance-\textbf{RE}educed Gradients) algorithm, 
which integrates sketch-based preconditioning to tackle ill-conditioning
and uses a scaled proximal mapping to minimize the non-smooth regularizer.
This stochastic variance-reduced algorithm achieves condition-number-free linear convergence to the optimum, 
delivering an efficient and scalable solution for 
ill-conditioned composite large-scale convex machine learning problems. 
Extensive experiments on lasso and logistic regression demonstrate that \sapph{} often converges $20$ times faster than other common choices such as \texttt{Catalyst}, \texttt{SAGA}, and \texttt{SVRG}.
This advantage persists even when the objective is non-convex or the preconditioner is infrequently updated, highlighting its robust and practical effectiveness. 

\end{abstract}
\section{Introduction} \label{intro}
Modern datasets in science and machine learning are massive in scale.
As an example in genetics, whole genome sequencing efforts on large-scale population cohorts like the Million Veterans Program, AllofUS program, and the OurFutureHealth project are expected to collect data from more than millions of individuals on billions of genetic variants.
Single-cell sequencing 
and epigenetic features such as DNA methylation levels, transcription factor binding, gene proximity, and other annotations can further increase the scale of the problem.
Naively training a machine learning model on such data leads to an expensive optimization problem whose solution is uninterpretable and often fails to generalize to unseen data.
Modern statistics and learning theory provide a solution to this challenge, 
by using \emph{structured regularization}
to improve model interpretability and generalization.
Mathematically, the optimization problem to solve is a regularized empirical risk minimization (rERM) problem,
\begin{equation}
\label{eq:model_prob}
\tag{rERM}
    \mathop{\text{minimize}}\limits_{w \in \R^p} \ \mathcal{R}(w) \assign \frac{1}{n}\sum_{i=1}^{n} \ell_i( w)+r(w),
\end{equation}
where $n$ is the number of samples, $p$ is the number of features, 
and $w \in \R^p$ represents the model weights.
Here the $\ell_i(w)$'s are smooth loss functions, and $r(w)$ is a possibly non-convex and non-smooth regularizer that encourages a parsimonious solution.
Popular regularizers include the $l_1$-norm, SCAD regularizer, or the indicator function for the $l_0$-ball. 
Problem \eqref{eq:model_prob} models many fundamental problems in machine learning, such as Lasso, elastic-net regression, $l_1$-logistic regression, dictionary learning, and matrix completion,
as well as modern applications such as convex neural networks \cite{pilanci2020neural, feng2024cronos}, data models for deep learning \cite{ilyas2022datamodels}, and pruned ensembles of trees \cite{liu2021controlburn}. \\

Realistic problems in high dimensions $n$ and $p$ are generally ill-conditioned, 
with a loss whose Hessian eigenvalues span many orders of magnitude \cite[Table 2]{frangella2024sketchysgd}.
Ill-conditioning requires first-order methods like stochastic gradient descent to use a small learning rate to avoid divergence, 
and hence to suffer from slow convergence.
For example, if $\ell(\cdot, w)$ is the loss of a generalized linear model (GLM), the conditioning of \eqref{eq:model_prob} is controlled by the conditioning of the data matrix $X$.
In large-scale datasets, the features are often highly correlated, so $X$ is approximately low-rank and has a large condition number---possibly larger than the sample size $n$, 
leading to a difficult optimization problem in \eqref{eq:model_prob}. 

\begin{figure}[h!]
        \centering
        \includegraphics[width=0.81\linewidth]{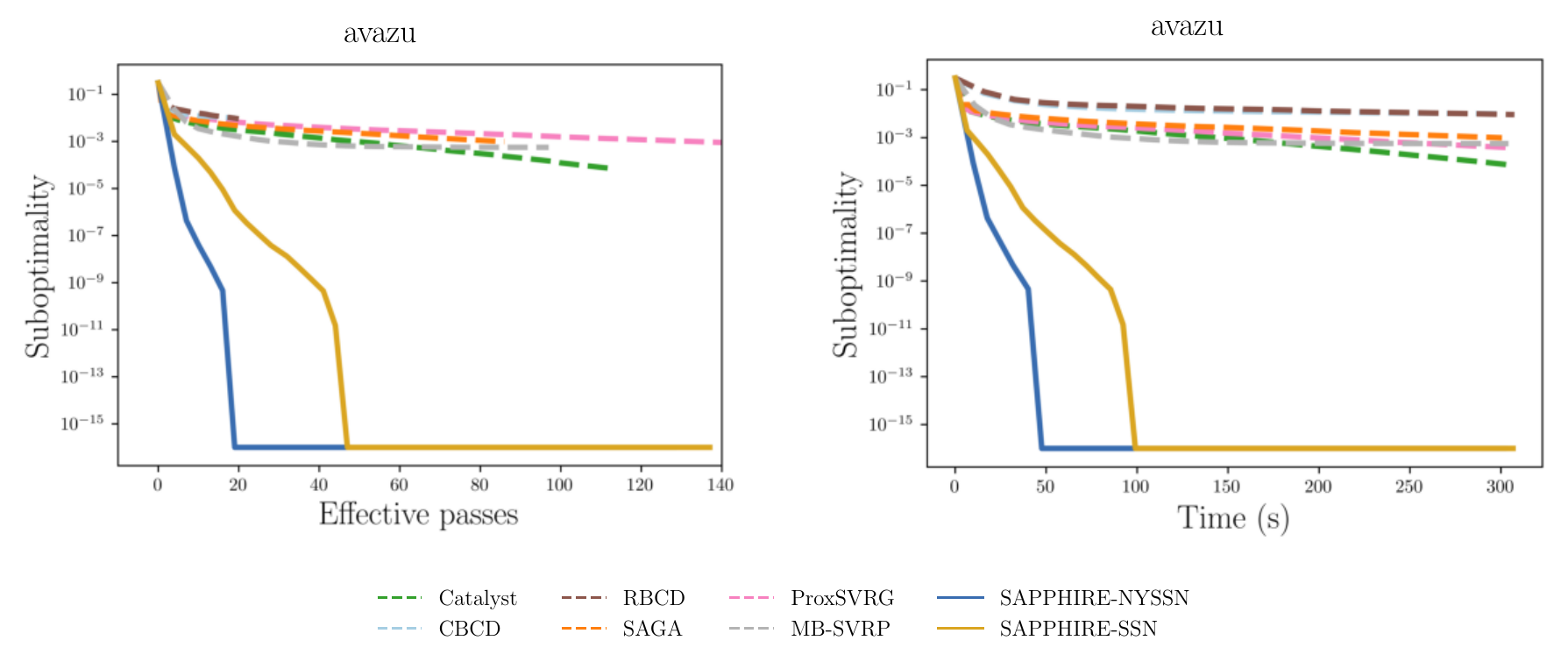}
        \caption{\sapph{} significantly outperforms competing stochastic optimizers on a large-scale click prediction problem with the avazu dataset $(n=12,642,186, \ p=999,990)$.}
        \label{fig:intro}
\end{figure}

A traditional way to mitigate ill-conditioning in optimization is to use second-order methods, such as Newton's method or BFGS, which incorporate curvature information. These methods are robust and can achieve local superlinear convergence.
While these classical methods do not scale to the big data regime, new stochastic second-order methods developed in the last decade can scale and deliver better practical performance than first-order methods \cite{byrd2011use, erdogdu2015convergence, pilanci2017newton, roosta2019sub, gower2019rsn, meng2020fast, promise}. Indeed, recent work \cite{promise} demonstrates that combining second-order information with variance-reduced gradients can yield fast stochastic second-order methods with strong theoretical and practical convergence. 
However, these methods work best for smooth and (strongly) convex problems, and cannot handle structured regularization with a non-smooth regularizer, such as the $\ell_1$ regularizer in the Lasso problem. \\

Structured regularization improves both interpretability and generalization. However, its effect on ill-conditioning is more nuanced. On one hand,  
near convergence,
the additional structure can help the algorithm identify a lower-dimensional basis for the solution and reduce the effective dimensionality of the problem. 
On the other hand, many structured penalties are non-smooth, which complicates algorithmic design and can worsen conditioning.
Thus, even with structured regularization, high-dimensional problems ($n,p\gg 1$) still suffer from ill-conditioning. \\

In this work, we address precisely these computational challenges, using stochastic second-order information to develop an efficient, scalable method that handles both non-smoothness and large-scale, ill-conditioned data.
The algorithm we design, \sapph{} (\textbf{S}ketching-based \textbf{A}pproximations for \textbf{P}roximal \textbf{P}reconditioning and \textbf{H}essian \textbf{I}nexactness with Variance-\textbf{RE}duced Gradients),
is a preconditioned variance-reduced stochastic gradient algorithm that generalizes the approach in \cite{promise} 
to the (non-smooth) regularized problem \eqref{eq:model_prob}.
\cref{fig:intro} shows the performance of \sapph{} with two different preconditioners on a large-scale (and hence ill-conditioned) logistic regression problem with an elastic-net penalty.
With either preconditioner, \sapph{} converges significantly faster than competing methods, demonstrating its robustness and efficiency. 


\subsection{Contributions}

We summarize our contributions as follows: 
\begin{enumerate}
    \item We introduce a robust framework, \sapph{}, to solve ill-conditioned composite large-scale convex optimization problems using variance reduction that requires only stochastic gradients and stochastic Hessians, and prove convergence of this framework under lazy preconditioner updates.
    \item \sapph{} accesses the non-smooth regularizer through a scaled proximal mapping in the preconditioned norm. While this mapping does not have a closed form, we propose to solve it iteratively using accelerated proximal gradient (APG) algorithm and demonstrate that only a few APG iterations are required.
    \item We prove that \sapph{} achieves global linear convergence for strongly convex objectives and global sublinear convergence for convex objectives. 
    We also show that the algorithm converges locally
    at a linear rate that is independent of the condition number. 
    Our theoretical analysis supports selection of hyperparameters for \sapph{}
    that lead to robust good performance without further data-dependent tuning. 
    \item Through large-scale experiments across diverse datasets,
    we demonstrate that \sapph{} often converges over $20$ times faster than other popular stochastic optimizers,
    both for convex and non-convex regularizers.
\end{enumerate}

\subsection{Roadmap}

We organize the paper as follows. Section \ref{related_work} reviews recent literature, highlighting connections to existing methods and the distinctions of our proposed algorithm. Section \ref{sec:alg} proposes the \sapph{} algorithm formally and elaborates on its core components of sketch-based preconditioning and scaled proximal mapping. Section \ref{sec:theory} establishes comprehensive convergence results for \sapph{}, covering both global and local convergence with various convexity assumptions. Section \ref{sec:exp} demonstrates the superior performance of the algorithm over popular tuned stochastic optimizers through extensive numerical experiments. 

\subsection{Notation}

Throughout the paper, $\| \cdot \|$ denotes the Euclidean norm, and denote $\lVert \cdot \rVert_A$ as the matrix norm induced by matrix $A$, where $\lVert x \rVert_A = \sqrt{x^\top A x}$. For a positive definite matrix $A$, we write $A \succeq 0$. The Loewner order is denoted by $\preceq$, where $A \preceq B$ if the matrix $B - A \succeq 0$. Given a positive definite matrix $A \in \mathbb{R}^{p \times p}$, its eigenvalues in descending order are written as $\lambda_1(A) \geq \lambda_2(A) \geq \cdots \geq \lambda_p(A)$. The condition number of $A$ is defined as $\kappa(A) = \lambda_1(A) / \lambda_p(A)$. For any scalar $\beta > 0$, we define the effective dimension $d_{\text{eff}}^{\beta}(A) = \text{tr}(A (A + \beta I)^{-1})$, which provides a smoothed measure of eigenvalues greater than or equal to $\beta$.

\section{Related Work} \label{related_work}

Here we review prior work on stochastic second-order methods, with particular emphasis on those developed for convex optimization problems, which is the main focus of this paper.

\paragraph{Variance-reduced stochastic first-order methods for finite sum minimization.}
Due to the massive size of contemporary machine learning datasets, much of the research in the past decade has focused on developing efficient algorithms that only require a stochastic first-order oracle.
The most successful of these algorithms are those that employ \emph{variance reduction}, which results in the variance of the gradient approaching zero as the iterates near an optimum \cite{johnson2013accelerating}. 
This technique yields global sublinear and linear convergence when the objective is convex and strongly convex, respectively.
Popular variance-reduced optimizers include \texttt{SAGA} \cite{defazio2014saga}, \texttt{ProxSVRG} \cite{xiao2014proximal}, \texttt{Catalyst} \cite{lin2018catalyst}, and \texttt{Katyusha} \cite{allen2018katyusha}.
These algorithms are also popular in practice for solving the empirical risk minimization problem \eqref{eq:model_prob}.
Indeed, the popular software package \texttt{scikit-learn} employs \texttt{SAGA} as the default stochastic gradient-based solver for problems such as logistic regression.
In the non-convex case, convergence to approximate stationary points has been established for many variants of these algorithms \cite{allen2016variance, reddi2016stochastic, j2016proximal, paquette2018catalyst, allen2018natasha}.
The assumptions underlying these theoretical guarantees typically prescribe that these methods use a minimal learning rate that goes to zero with $n$.
However, in practice, these algorithms are often run with a fixed learning rate as though the objective were convex, as this yields better performance \cite{j2016proximal,paquette2018catalyst}.

\paragraph{Stochastic second-order methods for finite sum minimization.}
Stochastic first-order methods suffer in the face of ill-conditioning.
To address this limitation, many authors have worked on stochastic second-order algorithms capable of scaling to large-scale machine learning problems. 
We classify these schemes by their target problems and methods used to compute gradients and Hessian. We summarize these results in Table \ref{tab:stoch-2nd-order}. 
Some methods require exact gradients at every iteration; some require only stochastic gradients; and some (``snapshot'') require stochastic gradients and occasional exact gradients.
All methods in the table require only stochastic samples of the Hessian.
Many assume interpolation ($\inf_w \mathcal R(w)=0$) to prove convergence to the global optimum.

\begin{table}[h]
\centering
\caption{Stochastic Second-Order Methods in ERM Literature
\label{tab:stoch-2nd-order}}
\resizebox{0.95\textwidth}{!}{
\begin{tabular}{cccccc}
\toprule
\textbf{Papers} & \textbf{Loss} & \textbf{Regularizer} & \textbf{Gradient} & \textbf{Fixed batchsize} & \textbf{Interpolation} \\
\midrule
\cite{byrd2011use, erdogdu2015convergence, berahas2016multi, pilanci2017newton, gower2019rsn, ye2021Appx, chen2022san, yuan2022sketched} 
& Convex
& None
& Exact 
& No
& No \\
\midrule
\cite{moritz2016linearly, bollapragada2018progressive, roosta2019sub, bollapragada2019exact, meng2020fast}
& Convex
& None
& Stochastic 
& No
& Yes \\
\midrule
\cite{derezinski2023stochastic, frangella2024sketchysgd, promise, garg2024second, mbsvrp, li2020subsampled}
& Strongly convex
& Smooth
& Snapshot 
& Yes 
& No \\
\midrule
\textbf{This paper}
& Convex
& Non-smooth
& Snapshot
& Yes
& No \\
\bottomrule
\end{tabular}
}
\end{table}

These work vary in how they use second-order information: 
some directly apply the inverse of the subsampled Hessian to the stochastic gradient \cite{roosta2019sub,bollapragada2019exact,meng2020fast}, or they use 
the subsampled Hessian-vector product to update the preconditioner rather than using the difference between two stochastic gradients \cite{moritz2016linearly,bollapragada2018progressive,meng2020fast}.
However, the theory underlying these methods requires large or growing gradient batch sizes \cite{roosta2019sub,bollapragada2018progressive,bollapragada2019exact}, periodic full gradient computation \cite{moritz2016linearly}, or interpolation \cite{meng2020fast}, which are unrealistic assumptions for large-scale convex problems.
Further, many of these methods lack practical guidelines for setting hyperparameters such as batch sizes and learning rate, leading to the same tuning issues that plague stochastic first-order methods. \\

Recently work has developed more practical stochastic second-order algorithms 
that use variance-reduction and stochastic second-order information to improve convergence \cite{derezinski2023stochastic, frangella2024sketchysgd, promise, garg2024second}.
The \texttt{PROMISE} framework in \cite{promise} leads to globally linearly convergent algorithms with \emph{constant} gradient batchsizes
and comes with theoretically-motivated default hyperparameter setting
that outperform tuned stochastic first-order methods. \\

However, most of these improved algorithms still assume smoothness and strong convexity to show their convergence results. For instance, \texttt{SVRN} \cite{derezinski2023stochastic, garg2024second} and \texttt{PROMISE} \cite{promise} require smooth and strongly convex objectives. \texttt{SketchySGD} \cite{frangella2024sketchysgd} can be used in the convex case but only converges to a noise ball around the optimum. \cite{mbsvrp} and \cite{li2020subsampled} can handle composite problems with a non-smooth regularizer in practice, but their convergence analyses are restricted to smooth and strongly convex problems. 
Therefore, \sapph{} fills a significant gap in the literature 
by providing condition-number-free linear convergence 
on convex composite problems \eqref{eq:model_prob}.\\

We will see in \cref{sec:exp} that \sapph{} often performs well on non-convex regularizers,
despite the lack of a convergence theory.
Provably convergent stochastic second-order methods for smooth non-convex finite sum minimization generally come in one of several flavors.
They use randomized approximation to the Hessian (via subsampling or sketching) together with 
cubic regularization \cite{kohler2017sub,tripuraneni2018stochastic,xu2020newton}, Newton-CG \cite{yao2023inexact, rathore2024challenges}, or trust region methods \cite{blanchet2019convergence, yao2021inexact, roosta2022newton} to (for example) guarantee convergence to a local minimum.
However, many of these methods require solving a challenging subproblem at each iteration, such as a cubic Newton step or a trust-region problem.
Consequently, these methods are often slower than stochastic first-order methods despite converging in fewer iterations.
\sapph{} lacks convergence guarantees for this setting, 
but offers a more palatable runtime. Combining an efficient method like \sapph{} with a safeguard to ensure convergence to a local minimum is an interesting challenge for future work. 

\subsection{Comparison with \texttt{SAPPHIRE}}

\cref{t-2ndOrdComp} positions \sapph{} relative to existing work on state-of-the-art stochastic second-order optimizers for solving instances of \eqref{eq:model_prob} with a loss that depends only on the inner product of the parameters and the data,
a model class that includes all (regularized) generalized linear models
\begin{equation}
\label{eq:l2_reg_glm}
\frac{1}{n}\sum_{i=1}^{n}\ell(a_i^{T}w)+\frac{\nu}{2}\|w\|^2+r(w).
\end{equation}
A restriction to $L_2$-regularized GLMs makes comparison to previous work as straightforward as possible, as \texttt{MB-SVRP}, \texttt{PROMISE}, and \texttt{Proximal Subsampled Newton} restrict their analysis to GLMs.
The table compares methods based on the properties they require to achieve condition number-free local convergence\footnote{We compare based on local and not global convergence as global convergence analyses are often looser and sometimes absent from previous work.}.
\cref{t-2ndOrdComp} considers whether the method allows for a non-trivial convex regularizer $r(w)$, its required gradient batchsize, and the size of the neighborhood of local convergence. 

\begin{table}[h]
\centering
\caption{\sapph{} vs. State-of-the-art competitors for solving \eqref{eq:l2_reg_glm}. 
Of the methods in the table, \sapph{} is the only variance-reduced stochastic gradient algorithm whose local convergence guarantees allow for a non-trivial convex regularizer. 
\sapph{} also has the best gradient batchsize requirement without requiring a smaller neighborhood of local convergence. 
\label{t-2ndOrdComp}}
\resizebox{0.85\textwidth}{!}{
\begin{tabular}{cccc}
\toprule
\textbf{Method} & \textbf{Regularizer} & \textbf{Gradient Batchsize} & \makecell{\textbf{Radius of} \\ \textbf{Local Convergence}} \\ 
\midrule
\texttt{SAPPHIRE} (\cref{alg:sapphire}) & Convex and Proxable & $\bigOt\left(\tau_\star^{\nu}\right)$ & $\bigO\left(\frac{\nu^{3/2}}{M}\right)$ \\
\midrule
\texttt{Proximal SSN} \cite{li2020subsampled} & Convex and Proxable & $n$ & $\bigO\left(\frac{\nu}{M}\right)$ \\ 
\midrule
\texttt{MB-SVRP} \cite{mbsvrp} & None & $\bigO\left(\chi^{\nu}(H(w_\star))\deff^{\nu}(H(w_\star))\kappa^{1/3}_{\textup{max}}\right)$ & $\bigO\left(\frac{\nu^4}{L^2_{\textrm{max}}M}\right)$ \\ 
\midrule
\texttt{SVRN} \cite{derezinski2023stochastic, garg2024second} & None & $\bigOt(\kappa_{\textup{max}})$ & $\bigO\left(\frac{\nu^{3/2}}{M}\right)$ \\
\midrule
\texttt{SketchySVRG} \cite{promise} & None & $\bigOt\left(\tau_{\star}^{\nu}\right)$ & $\bigO\left(\frac{\nu^{3/2}}{M}\right)$ \\
\bottomrule
\end{tabular}
}
\end{table}

\section{\sapph{}: A Fast Algorithm for Large-Scale Statistical Learning} 
\label{sec:alg}
In this section, we formally introduce the \sapph{} algorithm. 

\begin{algorithm}[h!]
\caption{\sapph{}
\label{alg:sapphire}}
    \KwIn{starting point $w_0$, gradient and Hessian batch $S_H, S_g$ with size $b_H$, $b_g$, preconditioner $P$, \\ \hspace{31 pt} preconditioner update times $\mathcal{U}$, learning rate multiplier $\alpha$, snapshot update frequency $m$}
    
    Initialize: snapshot $\tilde{w} = \tilde{w}_0$
    
    \For{$s = $ \rm{$0, 1, \ldots$ }}{

    Compute full gradient $\bar g = \nabla L(\tilde w)$

    Set $w_0 = \tilde{w}$

    \For{$k = $ \rm{$0, 1, \ldots m-1$ }}{
    
    \If{$ms + k \in \mathcal{U}$}{
        \rm{Sample batch $S_H$ to obtain indices for $\widehat \nabla^2 L(w_u)$}
        
        Compute preconditioner $P_k^{(s)}$: SSN \eqref{eq:ssn} or NySSN \eqref{eq:nyssn} with $\widehat \nabla^2 L(w_u)$
        \hfill $\vartriangleright$ Update preconditioner
    }
    Sample stochastic gradient batch $S_g$

    Compute estimator $\widehat{\nabla} L(w^{(s)}_k) = \frac{1}{b_g} \sum_{i \in S_g} \nabla \ell_i(w^{(s)}_k)$ and $\widehat{\nabla} L(\tilde{w}) = \frac{1}{b_g} \sum_{i \in S_g} \nabla \ell_i(\tilde{w})$
    
    Compute $v^{(s)}_k = \widehat{\nabla} L(w^{(s)}_k) - \widehat{\nabla} L(\tilde w) + \bar g$

    $w^{(s)}_{k+1} = \textbf{prox}_{\eta r}^{P_k^{(s)}} (w^{(s)}_k - \eta (P_k^{(s)})^{-1}v^{(s)}_k)$
    \hfill $\vartriangleright$ Evaluate scaled proximal mapping (Apply \cref{alg:apg})
    }
    Option 1:
    $\tilde{w} = \frac{1}{m} \sum_{k=1}^m w^{(s)}_k$
    \hfill $\vartriangleright$ Update snapshot as average of inner iterates
    
    Option 2:
    $\tilde{w} = w^{(s)}_m$
    \hfill $\vartriangleright$ Update snapshot as last iterate 
    }
\end{algorithm}

\subsection{\sapph{} algorithm}

\sapph{} is a preconditioned variance-reduced stochastic gradient algorithm based on the classic \texttt{ProxSVRG} algorithm from \cite{xiao2014proximal}. The most significant innovation of \sapph{} is the design of an effective preconditioner for the problem.
Preconditioning is critical to problems with large-scale data, often improving the runtime by orders of magnitude. However, preconditioning complicates the computation of the proximal operator. \\

In the following sections, we discuss how to construct the preconditioner, efficiently solve the associated scaled proximal mapping, and set algorithmic hyperparameters.


\subsection{Efficient preconditioning}

Preconditioning is a powerful technique to accelerate the convergence of optimization algorithms on ill-conditioned problems. A good preconditioner must effectively approximate the local Hessian while being fast to compute and to invert. \\

Classic methods from optimization, like Newton's method and BFGS, precondition the gradient using the (approximate) inverse Hessian.
As a result, these methods enjoy fast local convergence rates that are independent of the condition number. 
Unfortunately, the Hessian or Hessian approximation used by these methods is expensive to compute and to invert for large-scale problems.
Unfortunately, these methods fail to scale to the problems commonly encountered in machine learning.
Recent work \cite{erdogdu2015convergence,roosta2019sub,promise} has shown in the smooth non-composite, effective preconditioners can be constructed only using a small fraction of the data, reducing the cost of preconditioning substantially. \sapph{} adopts the Subsampled Newton and the Nystr{\"o}m Subsampled Newton preconditioners, motivated by the authors' prior work \cite{promise}.

\subsubsection{Subsampled Newton Preconditioner}
The subsampled Newton (SSN) preconditioner first introduced in \cite{roosta2019sub}, approximates the Hessian matrix $H_L(w) \in \mathbb{R}^{p \times p}$ of the smooth part of the objective in \eqref{eq:model_prob} using a subset $S_H \subset \{1,\ldots,n\}$ of the data with batch size $b_H = |S_H|$. 
The preconditioner is constructed as 
\begin{align} \label{eq:ssn}
    P = \frac{1}{b_H} \sum_{i \in S_H} \nabla^2 {\ell}_i(w) + \rho I,
\end{align}
where $\rho > 0$ is a regularization parameter that mitigates 
noise in the smaller eigenvalues of this preconditioner. \\

By using only a subset of the data, this approach significantly reduces computational cost compared to a full computation of the Hessian (as in Newton's method), yet still identifies essential information about the local curvature.
To understand the approximation qualities of the SSN preconditioner, we first recall the notion of $\rho$-Hessian dissimilarity from \cite{frangella2024sketchysgd}.

\begin{definition} Let $L(w)$ be as in \cref{eq:model_prob}, where each $\ell_i: \R^p\mapsto \R$ is a smooth convex function. Let $\rho\geq 0$ and $w_\in \R^p$, then for $\rho$-Hessian dissimilarity at $w$ is given by
\[
\tau^{\rho}(\nabla^2 L(w)) = \max_{i\in[n]} \lmax\left((\nabla^2 L(w)+\rho I)^{-1/2}(\nabla^2 \ell_i(w)+\rho I)(\nabla^2 L(w)+\rho I)^{-1/2}\right).
\]
Moreover, given a subset $\mathcal S$ of $\R^p$, we define the $\rho$-maximal Hessian dissimilarity over $\mathcal S$ by:
\[
\tau_\star^{\rho}(\mathcal S) = \sup_{w\in \mathcal S}\tau^{\rho}(\nabla^2 L(w)).
\]
\end{definition}
\begin{rem}
    When $\mathcal S = \R^p$, we will write $\tau_\star^{\rho}$ for shorthand. 
\end{rem}
$\rho$-Hessian dissimilarity measures how much an individual Hessian $\nabla^2 \ell_i(w)$ deviates from the average Hessian $\nabla^2 L(w)$.
When the $\nabla^2 \ell_i(w)$ are relatively similar to each other, the smaller $\tau^{\rho}(\nabla^2 L(w))$ is---in the extreme case all the $\nabla^2\ell_i(w)$ are the same, $\tau^{\rho}(\nabla^2 L(w)) = 1$. 
Conversely, when an outlier $\nabla^2\ell_i(w)$ exists, the dissimilarity can be as large as $n$. 
The following lemma from \cite{frangella2024sketchysgd} summarizes these facts.
\begin{lem}
    \label{lem:rho_dissim}
    For any $\rho \geq 0$ and $w\in \R^p$, the following inequalities holds
    \[
    \tau^{\rho}(w)\leq \min\left\{n, \frac{M(w)+\rho}{\mu+\rho}\right\},
    \]
    where $M(w) \coloneqq \lmax(\nabla^2 \ell_i(w))$.
    \[
    \tau_\star^{\rho} \leq \min\left\{n, \frac{L_{\textup{max}}+\rho}{\mu+\rho}\right\}.
    \]
\end{lem}
The $\rho$-Hessian dissimilarity can be far smaller than the upper bound in \cref{lem:rho_dissim} suggests. 
See \cite{frangella2024sketchysgd} for more details. 
This is significant as $\tau^{\rho}(\nabla^2 L(w))$ controls the sample size required to obtain a non-trivial approximation to the Hessian. 

\begin{lem}
    \label{lem:ssn_appx}
    Let $w \in \R^p$, $\zeta \in (0,1)$ and $\rho >0$. 
    Construct $\widehat \nabla^2 L(w)$ with $b_H = \bigO\left(\frac{\tau^{\rho}(\nabla^2L(w))}{\zeta^2}\log\left(\frac{\deff^{\rho}(\nabla^2L(w))}{\delta}\right)\right)$.
    Then, with probability at least $1-\delta$,
    \[
    (1-\zeta)P_{\textup{SSN}} \preceq \nabla^2 L(w)+\rho I \preceq (1+\zeta) P_{\textup{SSN}}.
    \]
\end{lem}

\subsubsection{Nystr{\"o}m Subsampled Newton Preconditioner} \label{pre:nyssn}

The second preconditioner \sapph{} utilizes is the Nystrom Subsampled Newton (NySSN) preconditioner introduced in \cite{frangella2024sketchysgd, promise}. 
The Nystr{\"o}m preconditioner computes a low-rank approximation of the Hessian matrix by projecting the subsampled Hessian onto a low-rank subspace in the span of $\Omega$. 
The Nystr{\"o}m Subsampled Newton preconditioner is given by
\begin{align} 
\label{eq:nyssn_naive}
    P = (\widehat \nabla^2 L(w) \Omega)(\Omega^\top \widehat \nabla^2 L(w)\Omega)^{-1} (\Omega^\top \widehat \nabla^2 L(w)) + \rho I
\end{align}
where $\Omega \in \mathbb{R}^{p \times r}$ is a random test matrix. 
Typical choices for $\Omega$ include standard normal random matrices, randomized trigonometric transforms, and sparse-sign matrices \cite{tropp2017fixed,frangella2023randomized}. \\

Constructing the NySSN preconditioner via \eqref{eq:nyssn_naive} is numerically unreliable due to the presence of the pseudoinverse.
Instead we apply the numerically stable procedure from \cite{tropp2017fixed} to compute the Nystr{\"o}m approximation: $(\widehat \nabla^2 L(w) \Omega)(\Omega^\top \widehat \nabla^2 L(w)\Omega)^{-1} (\Omega^\top \widehat \nabla^2 L(w))$.
The numerically stable procedure is presented in \cref{alg:nyssn} in \cite{randnys}.
It provides an approximate low-rank eigendecomposition of $\widehat \nabla^2 L(w)$: $\hat V \hat \Lambda \hat V^{\top}$.
Using the stable procedure, the NySSN preconditioner is given by
\begin{equation}
\label{eq:nyssn}
    P = \hat V \hat \Lambda \hat V^{\top} + \rho I.
\end{equation}
The preconditioner and its inverse can be applied to vectors in $\bigO(pr)$ time and requires $\bigO(pr)$ storage \cite{frangella2024sketchysgd, promise}. \\

This low-rank preconditioner is faster to invert for large-scale problems compared to the SSN preconditioner, especially when $b_H$ is large or the data is dense, and significantly reduces the computational cost of preconditioning.
Like the SSN preconditioner, the NySSN preconditioner admits strong theoretical guarantees. 
We have the following result from \cite{frangella2024sketchysgd}.

\paragraph{Theoretical guarantees}
\begin{lem}
    \label{lem:nyssn_appx}
    Let $w \in \R^p$, $\rho >0$, and $\gamma \geq 1$.
    Construct $H_{S_H}(w)$ with $b_H = \bigO\left(\tau^{\rho}(H(w))\log\left(\frac{\deff^{\rho}(H_{S_H}(w))}{\delta}\right)\right)$ samples and the Nystr{\"o}m approximation with rank $r = \bigO\left(\deff^{\gamma \rho}(H(w))+\log\left(\frac{1}{\delta}\right)\right)$.
    Then with probability at least $1-\delta$,
    \[
    \frac{1}{2\gamma}P_{\textup{NySSN}} \preceq H(w)+\rho I \preceq \frac{3}{2}P_{\textup{NySSN}}.
    \]
\end{lem}

\subsubsection{Preconditioner comparison}
We summarize the comparison between these two preconditioners in Table \ref{table:prec}. The SSN preconditioner  works best for sparse problems as it preserves the sparsity of data. In contrast, the NySSN preconditioner outperforms when the data is dense and has rapid spectral decay, so that a small rank $r$ suffices for an excellent approximation.

\begin{table}[h!]
    \caption{Comparison of Preconditioners
    \label{table:prec}}
    \centering
    \resizebox{0.8\textwidth}{!}{
    \begin{tabular}{c|c|c|c|c}
         \toprule
         & Construction Cost & Computation Cost & Memory Requirement & Best for \\ 
         \midrule
         SSN & $\mathcal{O}(b_H p + b_H^{3/2})$ & $\mathcal{O}(b_H p)$ & $\mathcal{O}(b_H p)$ & Sparse data \\
         NySSN & $\mathcal{O}(b_H rp)$ & $\mathcal{O}(rp)$ & $\mathcal{O}(rp)$ & Dense data \\
         \bottomrule
    \end{tabular}
    }
\end{table}

In \cref{sec:exp}, we demonstrate that NySSN performs at least as well as SSN, with especially strong performance on sparse data. 

\subsection{Scaled Proximal Mapping}
In contrast to \texttt{ProxSVRG}, to update the parameters, \sapph{} must evaluate the scaled proximal mapping:
\begin{align} \label{eq:scaled_prox}
    w_{k+1} = \textbf{prox}_{\eta r}^{P} (w_k - \eta P^{-1}v_k) 
            &:= \underset{{w\in \R^p}}{\mbox{argmin}} \left\{r(w)+\frac{1}{2\eta}\|w-P^{-1}\left(w_k-\eta v_k\right)\|_P^2\right\} \nonumber \\
    &= \underset{{w\in \R^p}}{\mbox{argmin}} \left\{\eta r(w)+ \langle \eta v_k, w-w_k\rangle +\frac{1}{2}\|w-w_k\|_{P}^2\right\}.
\end{align}
Unlike the traditional proximal operator, which often has a closed-form solution, \eqref{eq:scaled_prox} must be solved iteratively. 
For \sapph{} to be practical, it is essential that \eqref{eq:scaled_prox} be solved efficiently. 
\sapph{} uses the Accelerated Proximal Gradient (APG) algorithm \cite{beck2009fast, nesterov2013gradient} to solve \eqref{eq:scaled_prox}, motivated by three factors.
The first is that it is easy to apply the preconditioner $P$ to vectors, 
so computing the gradient of the smooth part of \eqref{eq:scaled_prox} is cheap.
The second is we can easily set the learning rate without resorting to line search---the smoothness constant is $\lambda_1(P)+\rho$, which is easy to compute for our preconditioners.
The third is that \eqref{eq:scaled_prox} is $\lambda_1(P)+\rho$-smooth and $\rho$-strongly convex and APG converges at the optimal rate of $\bigOt\left(\sqrt{\lambda_1(P)/\rho}\right)$.
We present pseudocode for APG applied to \eqref{eq:scaled_prox} in \cref{alg:apg}.


\begin{algorithm}[h!]
\caption{Accelerated Proximal Gradient (APG) for solving \eqref{eq:scaled_prox}.
\label{alg:apg}}
\KwIn{starting point $x_0$, preconditioner $P$, and regularization function $r$}
Initialize: $y_0 = x_0, s_0 = 1$

Set $\alpha = \left(\lambda_1(P)+\rho\right)^{-1}$

\For{$t = 0, 1, \ldots T$}{
    Calculate $x_{t+1} = \textbf{prox}_{\alpha\eta r}\left(y_t - \alpha(\eta v_t+P(x_t-w_k) \right)$
    
    Set $s_{t+1} = \frac{1}{2}(1 + \sqrt{1 + 4s_t^2})$
    
    Update $y_{t+1} = x_{t+1} + \frac{s_t - 1}{s_{t+1}} (x_{t+1} - x_t)$
}
\end{algorithm}




\section{Theory} 
\label{sec:theory}
In this section, we provide a convergence analysis for \sapph{}.
Our analysis shows \sapph{} converges to the global optimum linearly when $L(w)$ is smooth and $\Rc(w)$ is strongly convex, and sublinearly when $L(w)$ is smooth and $\Rc(w)$ is convex. 
We then provide concrete examples that illustrate when preconditioning improves convergence. 
In particular, when $L(w)$ is smooth and $\Rc(w)$ is strongly convex, 
we establish that \sapph{} enjoys local condition-number free convergence. 

\subsection{Quadratic Regularity}
We begin by recalling the definition of quadratic regularity, introduced in \cite{promise}.
\begin{definition}[Quadratic Regularity]
Let $f:\mathcal C \mapsto \R$ be a smooth convex function, where $\mathcal C$ is a closed convex subset of $\R^p$. The function $f$ is \emph{quadratically regular} if there exist constants $0<\gamma_\ell \leq \gamma_u <\infty$ such that for all $w_0, w_1, w_2 \in \R^p$,
\begin{equation} \label{def:quad_reg}
        \frac{\gamma_l(\mathcal C)}{2} \| w_2 - w_1 \|_{\nabla^2 f(w_0)}^2
        \leq f(w_2)-f(w_1) -\langle \nabla f(w_1), w_2 - w_1 \rangle 
        \leq \frac{\gamma_u(\mathcal C)}{2} \|w_2 - w_1\|_{\nabla^2 f(w_0)}^2. \\
\end{equation}
Here, $\gamma_u(\mathcal C)$ and $\gamma_l(\mathcal C)$ are called the \emph{upper and lower quadratic regularity constants}, respectively.
Moreover, if $f(w) = \frac{1}{n}\sum_{i=1}^{n}f_i(w)$ and each $f_i$ are $(\gamma_{u_i}, \gamma_{l_i})$-quadratically regular, we define
\[
\gamma_{u_{\textup{max}}}(\mathcal C) = \max_{i\in [n]} \gamma_{u_i}(\mathcal C), \quad \gamma_{l_{\textup{min}}}(\mathcal C) = \min_{i\in [n]} \gamma_{l_i}(\mathcal C).
\]
We also define the \emph{quadratic regularity ratio} and the \emph{maximal quadratic regularity ratio} as
\[
\q(\mathcal C) \coloneqq \frac{\gamma_u(\mathcal C)}{\gamma_l(\mathcal C)}, \quad \qmax \coloneqq \frac{\gamma_{u_{\textup{max}}}(\mathcal C)}{\gamma_l(\mathcal C)}.
\]
\end{definition}
\begin{rem}
    If $\mathcal C = \R^p$, we will omit explicitly writing $\mathcal C$ when presenting the quadratic regularity constants/ratios. 
\end{rem}

Quadratic regularity generalizes the traditional assumptions of smoothness and strong convexity to the Hessian norm. This assumption is critical to show convergence under infrequent preconditioner updates, as it allows $f$ to be upper and lower bounded in terms of the Hessian evaluated at where the preconditioner was constructed. Most importantly, quadratic regularity holds whenever the function in question is smooth and strongly convex.

\begin{lem}[Smoothness and strong-convexity imply quadratic regularity]
\label{lem:str_cvx_quad_reg}
    Let $f:\mathcal C \mapsto \R$ be a $\beta$-smooth $\mu$-strongly convex function, where $\mathcal C$ is a closed convex subset of $\R^p$. Then, $f$ is quadratically regular.
\end{lem}
Unfortunately, when $f$ is only smooth and convex, quadratic regularity fails: the Hessian is only guaranteed to be psd, and where it has a nullspace it cannot define a norm.
Instead, in this case, our convergence analysis rests on the weaker notion of $\rho$-\emph{weak quadratic regularity}.
\begin{definition}[$\rho$-weak quadratic regularity]
    Let $f:\mathcal C \mapsto \R$ be a smooth convex function, where $\mathcal C$ is a closed convex subset of $\R^p$. Then $f$ is $\rho$-weakly quadratically regular if the regularized function
    \[
    f_{\rho}(w) = f(w)+\frac{\rho}{2}\|w\|^2~ \textup{is quadratically regular.}
    \]
    We denote the corresponding quadratic regularity constants by: $\gamma^{\rho}_u$, $\gamma^{\rho}_l$, $\gamma^{\rho}_{u_{\textup{max}}},$ and $\gamma^{\rho}_{l_{\textup{min}}}$.
\end{definition}
We immediately conclude the following result from this definition and \cref{lem:str_cvx_quad_reg}.
\begin{lem}[Smoothness and convexity imply $\rho$-weak quadratic regularity] 
\label{lem:wk_quad_reg}
    If $f$ is $\beta$-smooth and convex, then it is $\rho$-weakly quadratically regular for any $\rho>0$.
\end{lem}
\paragraph{Three different scenarios.}
When analyzing \eqref{eq:model_prob} under the hypothesis of convexity, the standard regularity assumptions are: 1. The $\ell_i(w)$ are smooth and strongly convex for all $i\in [n]$ 2. The $\ell_i$ are smooth for all $i\in [n]$ and $L(w)$ is strongly convex, and 3. The $\ell_i(w)$ are smooth for all $i\in [n]$.
\cref{lem:str_cvx_quad_reg} and \cref{lem:wk_quad_reg} show these assumptions can be expressed in the language of quadratic regularity:
\begin{enumerate}[label=\arabic*)]
    \item $\ell_i(w)$ is $\beta_i$-smooth and strongly convex for all $i\in [n]$ $\implies \ell_i(w)$ is quadratically regular for all $\i\in[n]$ and $L(w)$ is quadratically regular.
    \item $\ell_i(w)$ is $\beta_i$-smooth and convex for all $i\in [n]$ and $L(w)$ is strongly convex $\implies \ell_i(w)$ is $\rho$-weakly quadratically regular for all $\i\in[n]$ and $L(w)$ is quadratically regular.
    \item $\ell_i(w)$ is $\beta_i$-smooth and convex for all $i\in [n]$ $\implies \ell_i(w)$ is $\rho$-weakly quadratically regular for all $\i\in[n]$ and $L(w)$ is $\rho$-weakly quadratically regular.
\end{enumerate}
Our analysis focuses on settings $1)$ and $3)$, as setting $2)$ is identical to setting $1)$ except for a change in one constant. 
We will elaborate on this point more below. 

\subsubsection{When does quadratic regularity improve over the condition number?}
\label{subsubsec:quad_reg_imp}
In this subsection, we provide intuition for the quadratic regularity ratio through examples that contrast it with the condition number, the quantity that typically appears in the analysis of optimization algorithms.
This discussion expands on that of \cite{promise}.
As our analysis depends on the quadratic regularity ratio and not the condition number,
our upper bounds are correspondingly tighter when the quadratic regularity ratio is smaller than the condition number.

\paragraph{Least-squares loss.}
Let $L(w) = \frac{1}{2n}\|A w - y\|^2 +\frac{\nu\|w\|_2^2}{2}$, where $A\in \R^{n\times p}$ and $\nu\geq 0$.
Since $L$ is a sum of quadratic functions, 
it has a constant Hessian and equals its own Taylor expansion.
It immediately follows that $\gamma_{l_i}=\gamma_{u_i}=1$.
Hence, $\q = \qmax = 1$. This ratio is much smaller than the condition number $\frac{\smax(A)^2+n\nu}{\smin(A)^2+n\nu}$ when the data matrix $A$ is ill-conditioned.

\paragraph{GLM on a bounded domain.}
A function $f$ is said to be $M$-quasi-self concordant ($M$-qsc) over $\mathcal C$ if
\[
D^3f(x)[u,u,v]\leq M\|u\|_{\nabla^2 f(x)}^2\|v\|\quad \forall x\in \mathcal C~\text{and}~\forall u,v \in \R^p, 
\]
where $D^3f(x)$ is the trilinear form representing the third derivative of $f$ \cite{nesterov2018lectures}.
{Let $R>0$ and suppose that $D=\textup{diam}(\mathcal C)\leq \log(R)/M$}.
Then the arguments of \cite{promise} show that
\[
\q(\mathcal C)\leq R^2, \quad \qmax(\mathcal C)\leq R^2. 
\]
Any GLM (which includes non-quadratic problems like logistic and Poisson regression) with a data matrix $A$ whose rows satisfy $\|a_i\|\leq 1$\footnote{This is a standard normalization step employed in packages like \texttt{scikit-learn} for stochastic optimizers like SAGA.} for all $i\in [n]$ is $1$-quasi-self-concordant \cite{karimireddy2018global, doikov2023minimizing}.
Thus, for $R = \mathrm{e}$, we have $\q(\mathcal C)\leq 8$. In contrast, the condition number of $L$ over $\mathcal C$ behaves like: $\kappa_L(\mathcal C) = \Theta\left(\frac{\sigma_{\mathrm{max}}^2(A)+n\nu}{\sigma_{\mathrm{min}}^2(A)+n\nu}\right)$, which is large when the data matrix $A$ is ill-conditioned.
This analysis shows that for objectives of interest, 
the quadratic regularity ratio may be a constant independent of the condition number
even when the function is not well approximated by a quadratic.

\subsection{Assumptions}
In this subsection, we introduce the additional assumptions in our analysis.
\begin{ass}[Convexity and smoothness]
\label{ass:cov}
    The non-smooth function $r(w)$ is lower semi-continuous and convex, and its effective domain dom$(r) = \{w \in \mathbb{R}^d \mid r(w) < +\infty\}$ is closed. 
\end{ass}
\cref{ass:cov} is standard and holds for all practical convex regularizers of interest.

\begin{ass}[$\zeta$-spectral approximation]
\label{ass:appr}
    There exists $\zeta \in (0, 1)$ such that for each $j \in \mathcal{U}$, 
    the preconditioner $P_j$ constructed at $w_j$ satisfies
   \[
   \begin{cases}
        (1 - \zeta) P_j \preceq \nabla^2 L(w_j) \preceq (1 + \zeta) P_j,
        & 
        \text{if $L(w)$ is quadratically regular,} \\
        \nabla^2 L(w_j) \leq (1 + \zeta) P_j
        &
        \text{if $L(w)$ is $\rho$-weakly quadratically regular.}
    \end{cases}
 \]   
\end{ass}

\cref{lem:ssn_appx} and \cref{lem:nyssn_appx} show that the SSN and NySSN preconditioners, when constructed properly, satisfy the conditions of \cref{ass:appr} with high probability.
Thus, \cref{ass:appr} can be viewed as conditioning on the good event that the appropriate approximation bound holds.
A similar assumption was made in \cite{promise}.
All our theorems can be shown to hold so long as \cref{ass:appr} holds with high probability: when the failure probability is sufficiently small, 
we can apply the law of total expectation to obtain the same rate with a slightly worse constant factor.
We rely instead on \cref{ass:appr} as it leads to simpler proofs
and allows us to establish the convergence of \sapph{} with any preconditioner that satisfies \cref{ass:appr}, 
rather than only for the SSN and NySSN preconditioners.

\subsection{Convergence of \sapph{}}
To establish convergence of \sapph{}, we must control the smoothness parameter of the stochastic gradient in the preconditioned norm in expectation. 
A constant $\Lc_P$ that provides an upper bound on this parameter is known as the \emph{preconditioned expected smoothness constant} \cite{promise, frangella2024sketchysgd}. 
The preconditioned expected smoothness generalizes the Euclidean norm-based expected smoothness constant from \cite{gower2019sgd} to preconditioned space. 
In the case when $r(w)=0$ in \eqref{eq:model_prob}, \cite{promise, frangella2024sketchysgd} have established bounds on the preconditioned expected smoothness constant. 
The following lemma provides an explicit expression for $\Lc_P$ in the general composite case. 
\begin{lem}[Preconditioned Expected Smoothness]
\label{lem:smooth}
    Instate \cref{ass:cov} and let each $\ell_i(w)$ in \eqref{eq:model_prob} be convex and twice-continuously differentiable. 
    Let $\rho>0$ and $P$ be a preconditioner constructed at $w_P\in \R^p$ satisfying  
    \[
    \nabla^2L (w_P) \preceq (1+\zeta) P.
    \]
    Then for any $w \in \R^p$, if each $\ell_i(w)$ in \eqref{eq:model_prob} is quadratically regular, then
    \[
        \E\|\widehat{\nabla} L(w) - \widehat{\nabla} L(w_\star) \|_{P^{-1}}^2 \leq 2 \mathcal{L_P} [\Rc(w) - \Rc(w_\star)],
    \]
    where
    \[
        \Lc_P = \left(\frac{n (b_g-1)}{b_g (n-1)} \gamma_u + \tau_\star^{\rho}\frac{n - b_g}{b_g (n-1)}\gamma_{u_{\textup{max}}}\right) (1 + \zeta).
    \]
\end{lem}
The proof is provided in \cref{subsec:lem_smooth_pf}.\\

\cref{lem:smooth} extends the classical smoothness condition in deterministic optimization to the stochastic and preconditioned setting and establishes a direct relationship between the preconditioned gradient norm variance and the suboptimality of $\Rc(w) - \Rc(w^\star)$. 
It generalizes the results of \cite{promise,frangella2024sketchysgd} to the convex composite setting. 
If the individual $\ell_i$'s are $\rho$-weakly quadratically regular, then $\mathcal{L_P}$ in \cref{lem:smooth} will be constructed by $\gamma^{\rho}_u, \tau_\star^{\rho}, \text{ and }\gamma^{\rho}_{u_{\textup{max}}}$.

\begin{lem}[Preconditioned Stochastic Variance]
\label{lem:var}
    Instate \cref{ass:cov} and \cref{ass:appr}, 
    and define the variance-reduced stochastic gradient at inner iteration $k$ in outer iteration $s$, $v_k^{(s)} = \widehat{\nabla} L(w_k^{(s)}) - \widehat{\nabla} L(\hat{w}^{(s)}) + \nabla L(\hat{w}^{(s)})$. The variance of this stochastic gradient is bounded in the preconditioned norm as
    \begin{align*}
        \mathbb{E} \| v_k^{(s)} - \nabla L(w_k^{(s)}) \|_{(P_k^{(s)})^{-1}}^2 \leq 4 \mathcal{L_P} [\Rc(w_k^{(s)}) - \Rc(w_\star) + \Rc(\hat{w}^{(s)}) - \Rc(w_\star)]. 
    \end{align*}
\end{lem}
The proof is provided in \cref{subsec:lem_var_pf}.\\

\cref{lem:var} shows that by employing the variance-reduced stochastic gradient $v_k^{(s)}$, we are guaranteed that the variance of the stochastic gradient goes to zero as we approach the optimum.
This property is essential to establishing convergence. If the gradient variance does not go to zero as we approach the optimum, we can only reach a neighborhood of the optimum with a fixed stepsize.  

\begin{subsubsection}{Convergence for quadratically regular $L$}
Here, we establish global convergence of \sapph{} under quadratic regularity of $L$.
For brevity, we only consider the case when each $\ell_i(w)$ is quadratically regular.
The argument and resulting statements for the case when the $\ell_i(w)$ are only $\rho$-weakly quadratically regular are identical, except that we replace $\Lc_P$ by $\Lc_{P_\rho}$.

\begin{thm}[Global Linear Convergence]
\label{thm:conv_str_cvx}
    Instate \cref{ass:cov} and \cref{ass:appr}.
    Suppose each $\ell_i(w)$ is quadratically regular. 
    Run \cref{alg:sapphire} with learning rate $0< \eta < \frac{1}{4\Lc_P}$. 
    Then the output of \cref{alg:sapphire} satisfies
    \[
        \E[\Rc(\hat w^{(s)})-\Rc(w_\star)] \leq \left(\frac{1}{(1-\zeta) \gamma_{\ell} \eta (1 - 4\eta\mathcal L_P)m} + \frac{4 \eta \mathcal L_P (m+1)}{(1 - 4\eta\mathcal L_P)m} \right)^{s} \left(\Rc(w_0) - \Rc(w_\star)\right).
    \]
    Thus, setting $\eta = \bigO(1/\Lc_{P})$ and $m = \bigO(\frac{\Lc_P}{(1-\zeta)\gamma_\ell})$, we have
    \begin{align*}
         \E[\Rc(\hat w^{(s)})-\Rc(w_\star)] \leq \left(\frac{2}{3}\right)^{s}\left(\Rc(w_0) - \Rc(w_\star)\right).
    \end{align*}
    Hence, the error falls below $\epsilon > 0$ after $s \geq 3 \log \left(\frac{\Rc(\hat{w}^{(0)}) - \Rc(w_\star)}{\epsilon} \right)$ outer iterations
    and the total number of stochastic gradient queries needed to reach an $\epsilon$-suboptimal point is bounded by
    \begin{align}
        \bigO \left( \left( n + \frac{n}{1-\zeta}\left(\frac{b_g-1}{n-1}\q+\frac{\tau^{\star}_\rho}{n}\frac{n-b_g}{n-1}\qmax\right)\right) \log \left(\frac{1}{\epsilon} \right) \right).
    \end{align}
\end{thm}
The proof of \cref{thm:conv_str_cvx} is provided in \cref{subsection:str_cvx_pf}. \\

\cref{thm:conv_str_cvx} establishes global linear convergence of \sapph{} when $L$ is quadratically regular and each $\ell_i$ is quadratically regular.
It substantially generalizes Theorem 17 in \cite{promise}, which only establishes convergence in the special case $r(w) = \nu/2\|w\|_2^2$.
In the preconditioned setting, the role of the condition numbers  $\kappa$ and $\kappa_{\textup{max}}$ are played by the quadratic regularity ratios $\q$ and $\qmax$. 
The convergence rate is controlled by a convex combination of $\q$ and $\qmax$, which captures the benefits of minibatching. 
As $b_g$ increases from $1$ to $n$, the  
weight on the smaller ratio $\q$ approaches unity, while the weight on $\qmax$ approaches $0$.
When $\q, \qmax = \bigO(1)$, which corresponds to the setting when preconditioning helps globally, the total number of gradient queries scales as
\[
    \bigO \left( \left( n + \frac{n}{1-\zeta}\right) \log \left(\frac{1}{\epsilon} \right) \right).
\]
Thus, \texttt{SAPPHIRE's} convergence rate is completely determined by the quality of the preconditioner, whose impact on the convergence rate comes through the $(1-\zeta)^{-1}$ factor. 
In the case when $1-\zeta = \Omega(1)$, \sapph{} exhibits the optimal number of queries $\bigO(n\log(1/\epsilon))$. \\

\begin{rem}
    If the regularizer corresponds to a projection onto a closed convex set $\mathcal C$, then $\q$ and $\qmax$ in \cref{thm:conv_str_cvx} should be replaced by $\q(\mathcal C)$ and $\qmax(\mathcal C)$.
\end{rem}

\cref{thm:conv_str_cvx} along with our discussion in \cref{subsubsec:quad_reg_imp} immediately yields the following corollary, which provides two concrete settings where \sapph{} exhibits an optimal convergence rate.
\begin{coro}
    Under the hypotheses of \cref{thm:conv_str_cvx} with the additional assumption that $1-\zeta = \Omega(1)$, the following statements hold:
    \begin{enumerate}
        \item Suppose $L(w) = \frac{1}{2n}\|Aw-b\|^2 + \frac{\nu\|w\|^2}{2}$ and $r(w) = \mu\|w\|_1$. 
        Run \cref{alg:sapphire} with $\mathcal U = \{0\}$, $\eta = \bigO(1)$, $m = \bigO(1)$ inner iterations, and $s=\bigO\left(\log\left(\frac{1}{\epsilon}\right)\right)$ outer iterations.
        Then \cref{alg:sapphire} converges to expected loss $\epsilon$ with the total number of full gradient queries bounded as $\bigO(n\log(1/\epsilon))$.
        \item Suppose $L(w) = \frac{1}{n}\sum_{i=1}^{n}\ell(a_i^{T}w)+ \frac{\nu\|w\|}{2}^2$, with $\|a_i\|\leq 1$ for all $i\in[n]$ and $r(w) = 1_{\mathcal C}$, where $\mathcal C$ is a closed convex set with $\textup{diam}(\mathcal C) \leq 2$. Run \cref{alg:sapphire} with $\mathcal U = \{0\}$, $\eta = \bigO(1)$, $m = \bigO(1)$ inner iterations, and $s=\bigO\left(\log\left(\frac{1}{\epsilon}\right)\right)$ outer iterations.
        Then converges to expected loss $\epsilon$ with the total number of full gradient queries bounded as $\bigO(n\log(1/\epsilon))$. 
    \end{enumerate}
\end{coro}
\end{subsubsection}

\begin{subsubsection}{Convergence for convex $\rho$-weak quadratically regular $L$}
When $L(w)$ is only convex and smooth, a common setting in large-scale machine learning problems, i.e., Lasso, \sapph{} admits the following ergodic convergence guarantee. 
\begin{thm}[\sapph{}: Convex $\rho$-Weak Quadratically Regular Convergence]
\label{thm:conv_cvx}
    Instate \cref{ass:cov} and \cref{ass:appr}. 
    Fix $m>0$.
    Suppose each $\ell_i(w)$ is convex and $\rho$-weakly quadratically regular.
    Run \cref{alg:sapphire} with Option 2. and learning rate $\eta = \min\{\frac{1}{4\mathcal L_P(m+2)}, \frac{1}{8(m+2)}\}$. 
    Then, after $S$ outer iterations,
    \[
    \E\left[\Rc\left(\frac{1}{Sm}\sum_{s=0}^{S-1}\sum_{k=1}^{m}\hat w_k^{(s)}\right) - \Rc(w_\star)\right]\leq \frac{48(\mathcal L_P^2+4)(m+2)}{S}\|w_0-w_\star\|_{P_0^{(0)}}^2+\frac{12(\mathcal L_P +2)}{S}\left(\Rc(w_0)-\Rc(w_\star)\right).
    \]
    Thus, after $S = \bigO\left(\frac{m \mathcal L_P^2}{\epsilon}\right)$ outer iterations,
    \[
    \E\left[\Rc\left(\frac{1}{Sm}\sum_{s=0}^{S-1}\sum_{k=1}^{m}\hat w_k^{(s)}\right) - \Rc(w_\star)\right] \leq \epsilon \left[\|w_0-w_\star\|_{P_0^{(0)}}^2+\Rc(w_0)-\Rc(w_\star)\right].
    \]
\end{thm}
The proof of \cref{thm:conv_cvx} is provided in \cref{subsection:conv_cvx_pf}.\\

\cref{thm:conv_cvx} establishes that \sapph{} converges ergodically at an $\bigO\left(1/\epsilon\right)$ rate, matching the rate of gradient descent in the smooth convex case and \texttt{ProxSVRG} without preconditioning \cite{poon2018local}. 
Unfortunately, the dependence of $S$ on $m$ in the theorem implies the total gradient queries scale as $\bigO(\frac{n+m^2\mathcal L_P^2}{\epsilon})$, rather than the expected $\bigO(n+\mathcal L_P/\epsilon)$. 
This coupling also appears in analysis without preconditioning \cite{poon2018local}, with a rate of $\bigO(\frac{n+m^2\mathcal L}{\epsilon})$, so this issue does not stem from \sapph{} employing preconditioning.
The issue could be avoided by combining \sapph{} with a black-box reduction such as \texttt{AdaptReg} \cite{allen2016optimal}, which is based upon approximately minimizing a sequence of strongly convex surrogates. 
However, we have not found this to be necessary in practice. 
The suboptimal dependence on $m$ arises because \cref{thm:conv_cvx} assumes the very conservative hyperparameter setting: $\eta = \bigO(1/(\mathcal L_P m))$.
In practice, we run \sapph{} with $\eta = \bigO(1/\mathcal L_P)$, which corresponds to the setting in \cref{thm:conv_str_cvx} when $L(w)$ is quadratically regular.
While this more aggressive hyperparameter setting is not supported by \cref{thm:conv_cvx}, it yields excellent empirical performance in practice (\cref{sec:exp}).
The theory-practice gap in the setting of $\eta$ shows \cref{thm:conv_cvx} is overly conservative in the requirements it stipulates for \sapph{} to converge. 

\paragraph{When global convergence rates are pessimistic.} 
\cref{thm:conv_cvx} can overestimate the time needed to solve \eqref{eq:model_prob} when the regularizer is structured. 
Consider the Lasso problem where $L(w) = \frac{1}{2n}\|Xw-y\|^2$, $X\in \R^{n\times p}$ with $p>n$, and  $r(w) = \lambda \|w\|_1$.
When $p>n$, the covariance matrix $\frac{1}{n}X^{T}X$ is degenerate, so $L(w)$ is convex but not strongly convex.   
However, the defining property of the Lasso model is that the solution vector $w_\star$ is sparse.
When restricted to the support set of the solution $w_\star$, the covariance matrix is often no longer degenerate, so strong convexity holds as long as the iterates stay on the support set, which implies a linear convergence rate.
Optimization algorithms that identify the low-dimensional manifold on which the solution lives within a finite number of iterations and remain there are said to possess the \emph{manifold identification property} \cite{liang2017activity,liang2017local,sun2019we}.
Variance-reduced stochastic gradient methods like \texttt{ProxSVRG}, \texttt{SAGA}, and \sapph{} possess this property \cite{poon2018local}.
Hence, for problems like the Lasso, \sapph{} will exhibit an initial sublinear convergence phase, followed by a linearly convergent phase once it has identified the manifold on which the solution lives.
For some problem instances, this identification occurs rapidly so that the linearly convergent phase dominates---in which case the rate predicted by \cref{thm:conv_cvx} is highly pessimistic.
The manifold identification property can still be beneficial even when the objective is globally strongly convex, as with the elastic net. 
On the low-dimensional manifold, $L(w)$ can be better conditioned than it is globally, so the preconditioner does not have to be as good to ensure the preconditioned condition number is close to unity. 

\end{subsubsection}
\begin{subsection}{Local Convergence of SAPPHIRE}
In this subsection, we establish the local condition number free convergence of \sapph{}.
We focus on the case that each $\ell_i(w)$ is $\nu$-strongly convex and has an $M$-Lipschitz Hessian. 
Local convergence is established within the following neighborhood of the optimum $w_\star$:
\[
\Nstar \coloneqq \left\{\|w-w_\star\|^2_{\nabla^2 L(w_\star)} \leq \frac{\nu^{3/2}}{2M} \right\}.
\]
The key to achieving fast local convergence is that within $\Nstar$, the quadratic regularity constants are guaranteed to be very close to unity, enabling us to establish the following result. 
\begin{thm}
\label{thm:sapph_loc_con}
    Let $\varepsilon_0 \in (0, 1/6]$.
    Suppose that each $\ell_i$ is $\nu$-strongly convex, and has an $M$-Lipschitz Hessian, and that $w_0\in \Nstar$.
    Instate \cref{ass:cov} and \cref{ass:appr} with $\zeta = \varepsilon_0$. 
    Run \cref{alg:sapphire} using Option 2 with $\mathcal U = \{0\}$, 
    $m = 10$ inner iterations, $s = 2\log\left(\frac{1}{\epsilon}\right)$ outer iterations, $\eta = 1$, and $b_g = \bigOt\left(\tau^{\rho}(\Nstar)\log(\frac{1}{\delta})\right)$. 
    Then, with probability at least $1-\delta$,
    \[
    \|\hat w^{(s)}-w_\star\|_{\nabla^2 L(w_\star)}\leq \epsilon.
    \]
    Hence, the total number of stochastic gradient queries within $\epsilon$ distance of the optimum is bounded by
        \[
        \bigOt\left(n\log\left(\frac{1}{\epsilon}\right)\right)
        \]
\end{thm}
The proof of \cref{thm:sapph_loc_con} is provided in \cref{sec:sapph_fast_local_convergence}. \\

\cref{thm:sapph_loc_con} shows that within in $\Nstar$, \sapph{} enjoys linear convergence independent of the condition number.
It provides a generalization of Theorem 19 in \cite{promise} to the strongly convex composite setting. 
As in \cite{promise}, the required gradient batchsize only scales as $\bigOt\left(\tau^{\nu}(\Nstar)\right)$, which is never larger than the condition number $\kappa$ or $n$ and is often significantly smaller, as we shall see shortly below when we specialize to GLMs.
Having a gradient batchsize requirement independent of $\kappa$ is crucial in the ill-conditioned setting common in large-scale machine learning, where we can easily have $\kappa>n$. \\

To make \cref{thm:sapph_loc_con} more concrete, we present the following corollary, which specializes to the case when $L(w)$ corresponds to a GLM. 

\begin{coro}
    \label{corr:glm}
    Let $A \in \R^{n\times p}$, and let $a_i\in \R^p$ denote the $i$th row of $A$.
    Under the hypotheses of \cref{thm:sapph_loc_con}, suppose that $\ell_i(w) = \ell(a_i^{\top}w)+\frac{\nu\|w\|^2}{2}$, $\frac{1}{n}\lambda_j(A^\top A) \leq Cj^{-2\beta}$ for $\beta>1$, and $\nabla^2 L(w_\star)$ is ridge-leverage incoherent.
    Then if $b_g = \bigO\left(\sqrt{n}\log\left(\frac{1}{\delta}\right)\right)$, it holds with probability at least $1-\delta$ that only
    \[
      \bigOt\left(n\log\left(\frac{1}{\epsilon}\right)\right),
    \]
    stochastic gradient evaluations are required to ensure the output of \cref{alg:sapphire} satisfies
    \[
    \|\hat w^{(s)}-w_\star\|_{\nabla^2 L(w_\star)}\leq \epsilon.
    \]
\end{coro}

\cref{corr:glm} shows that under a spectral decay condition on $A$ that commonly arises in machine learning problems, \sapph{} only needs to use a batchsize of $\bigOt\left(\sqrt{n}\right)$ to ensure a condition number-free convergence with high probability.
Thus, we can set $b_g$ to be far smaller than $n$, while ensuring a fast convergence rate.
This concrete example shows that the dependence upon $\tau_\star^{\rho}(\Nstar)$ yields real improvements over results where the batch size depends upon $\kappa$.

\end{subsection}

\section{Experiments} \label{sec:exp}

In this section, we verify the effectiveness of \sapph{} (\cref{alg:sapphire}) with experiments on real-world data on a variety of machine learning tasks.
Our experiments utilize a diverse collection of datasets, which capture a variety of settings: (big-data) $n\gg p$, wide-data ($p\gg n$), and big and high-dimensional ($n\sim p$).
We refer to datasets with size $n < 10^6$ to be medium-scale and $n \geq 10^6$ to be large-scale. 
Moreover, we consider datasets of varying degrees of sparsity, ranging from extremely sparse to completely dense. 
Thus, the datasets we use capture the variety of possible scenarios that occur in machine learning.
Detailed statistics are presented in \cref{table:data}. \\

To demonstrate the effectiveness of \sapph{} over existing optimizers, we compare compare with the following standard stochastic first-order algorithms for solving \eqref{eq:model_prob}: \texttt{Catalyst} \cite{lin2018catalyst}, Cyclic Block Coordinate Descent (\texttt{CBCD}) \cite{beck2013convergence, li2017cbcd}, \texttt{ProxSVRG} \cite{xiao2014proximal}, Randomized Block Coordinate Descent (\texttt{RBCD}) \cite{richtarik2014iteration, lu2015complexity}, and \texttt{SAGA} \cite{defazio2014saga}.
We also compare with the stochastic second-order method \texttt{MB-SVRP} \cite{mbsvrp} to show that \sapph{} outperforms existing stochastic second-order methods for solving \eqref{eq:model_prob}. 
We evaluate the performance of algorithms using two metrics: wall-clock time in seconds, and the number of effective passes which corresponds to the total number number of full gradient evaluations made by each method. 


\begin{table}[h!]
    \caption{Datasets Statistics Summary
    \label{table:data}}
    \centering
    \begin{tabular}{c|c|c|c|c}
         \toprule
         Dataset & Training samples & Test samples & Variables & Non-zeros \% \\ 
         \midrule
         rna-seq & 640 & 160 & 20530 & $85.83\%$ \\
         rcv1 & 20242 & 677399 & 47236 & $0.02\%$ \\
         p53 & 25136 & 6284 & 5408 & $98.51\%$ \\
         yearmsd & 463715 & 51630 & 90 & $100\%$ \\
         covtype & 464810 & 116201 & 54 & $22\%$ \\ 
         url & 1916904 & 479226 & 3231961 & $0.01\%$ \\
         avazu & 12642186 & 1719304 & 999990 & $0.01\%$ \\
         \bottomrule
    \end{tabular}
\end{table}

\paragraph{Overview of the experiments.} 
The next four subsections provide a detailed comparison of \sapph{} and competing methods on 
(\cref{sub: mc}) medium-scale convex problems, (\cref{sub: lc}) large-scale convex problems, (\cref{sub: mnc}) medium-scale non-convex problems, and (\cref{sub: lnc}) large-scale non-convex problems.

\subsection{Convergence on convex objectives}

This section describes experiments on convex objective functions $\mathcal{R}$, including logistic regression and least-square regression with the Lasso and elastic-net regularizers. 

\subsubsection{Medium-Scale Convex Problems}
\label{sub: mc}
\begin{paragraph}{$L_1$-Logistic regression and Lasso.}
Every method was given a maximum runtime of 120 seconds and a limit of 200 full-gradient evaluations. 
Results for the medium-scale experiments on $L_1$-Logistic regression and Lasso are presented in \cref{fig:medium-1} and \cref{fig:medium-2}.
Recall we are in the setting of \cref{thm:conv_cvx}, as $L(w)$ is only guaranteed to be smooth and convex.
\sapph{} yields the best performance across all tasks. 
Moreover, we observe an initial sublinear phase of convergence followed by linear convergence once \sapph{} identifies the support of the solution.
The observed convergence behavior is consistent with the guarantees of \cref{thm:conv_cvx}, verifying our theory. 
    \begin{figure}[t]
        \centering
        \includegraphics[scale=0.5]{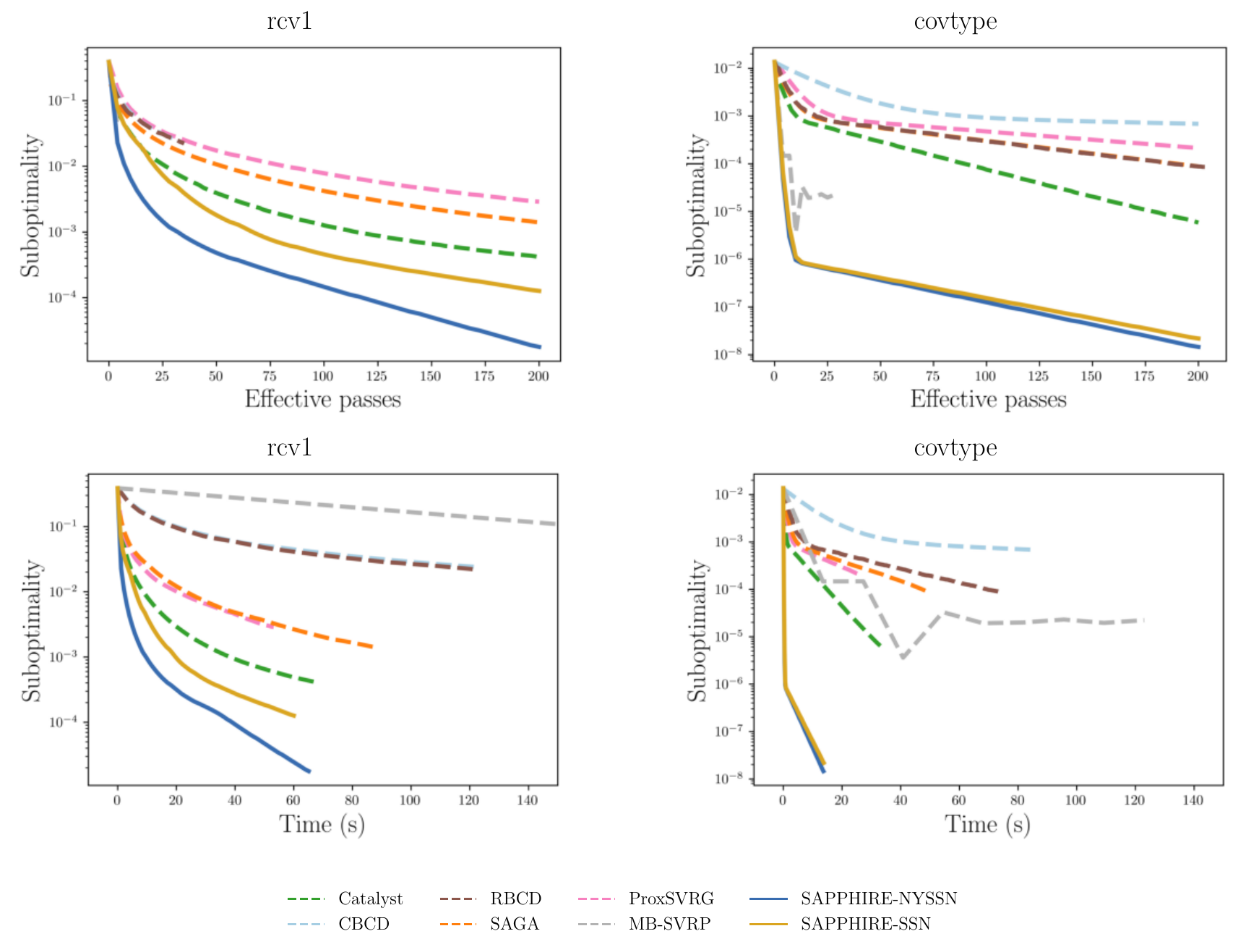}
        \caption{$L_1$-logistic regression: \sapph{} vs. competing methods on rcv1 and covtype}
        \label{fig:medium-1}
    \end{figure}

    \begin{figure}[t]
        \centering
        \includegraphics[scale=0.5]{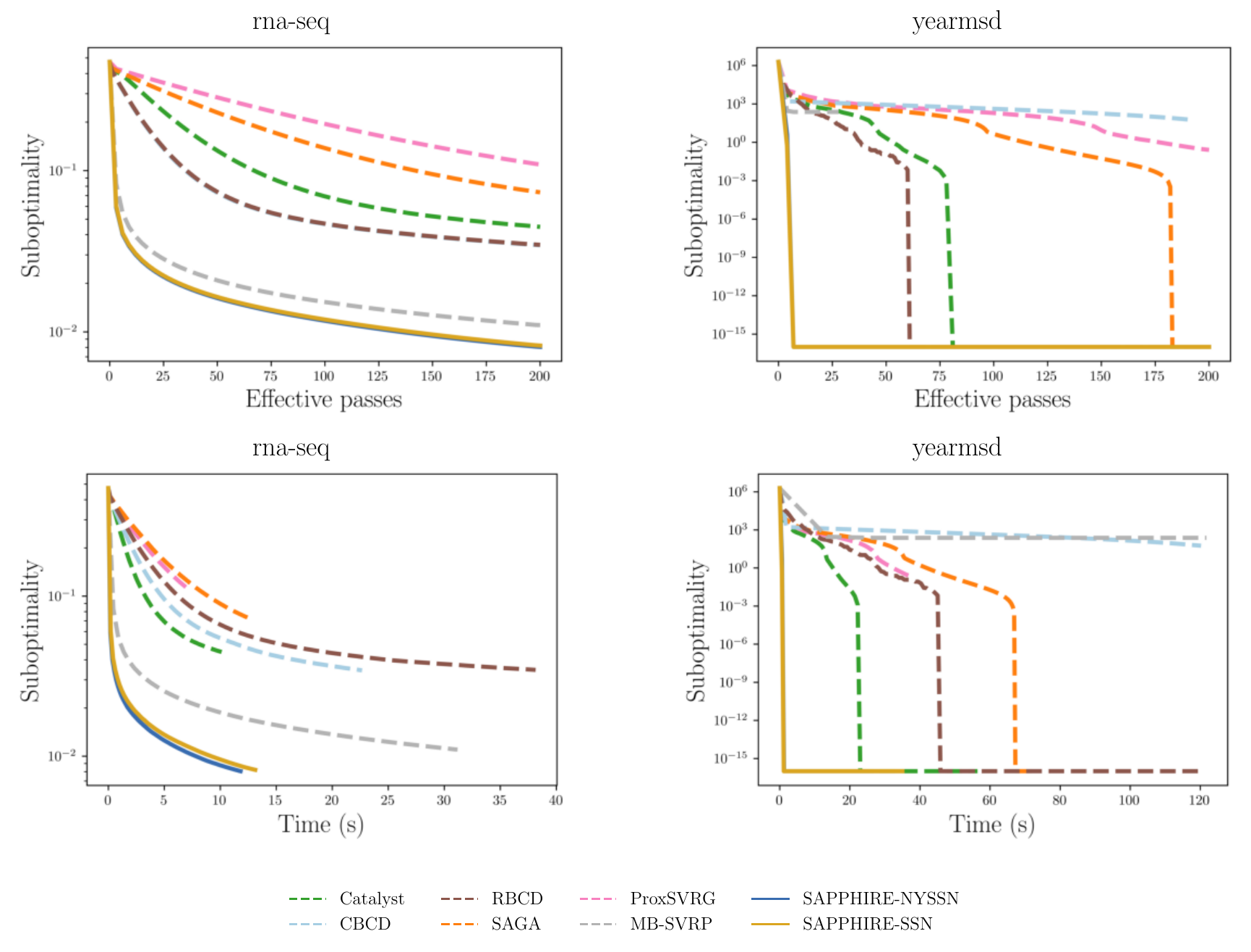}
        \caption{Lasso: \sapph{} vs. competing methods on rna-seq and yearmsd}
        \label{fig:medium-2}
    \end{figure}
\end{paragraph}

\subsubsection{Large-Scale Convex Problems} \label{sub: lc}

\begin{figure}[h!]
    \centering
    \includegraphics[scale=0.5]{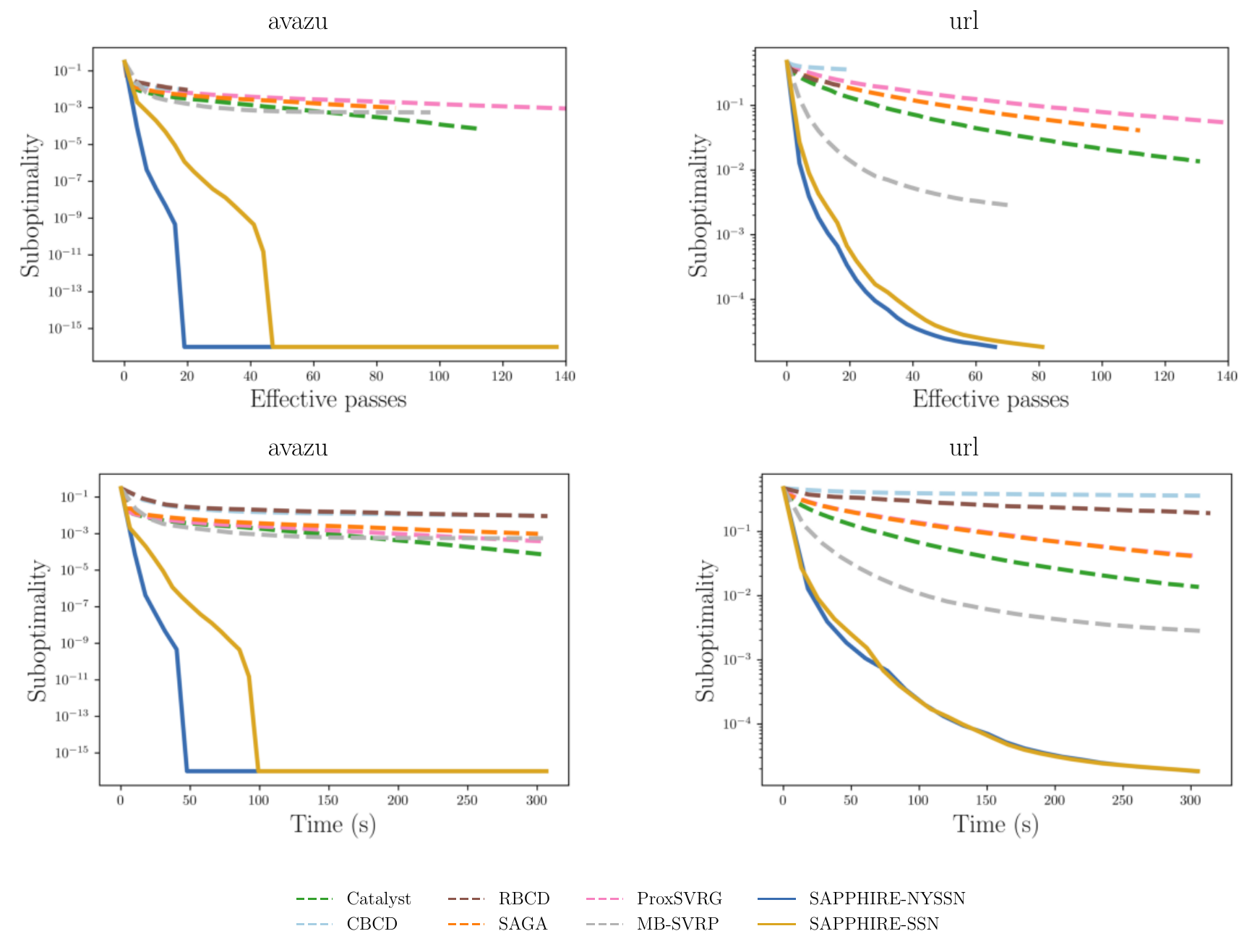}
    \caption{Logistic regression: \sapph{} vs. competing methods on avazu and url}
    \label{fig:large-1}
\end{figure}

\paragraph{Logistic regression with elastic-net penalty.}
In this part, we consider logistic regression with an elastic-net penalty on the large-scale datasets avazu and url.
Note, the elastic-net penalty ensures that this problem is strongly convex. 
Every method was given a maximum runtime of 300 seconds and a limit of 200 full-gradient evaluations. 
The results of these experiments are presented in \cref{fig:large-1}. \\

Similar to the medium-scale experiments, we observe that \sapph{} exhibits the best performance on both datasets. 
On avazu \sapph{} exhibits dramatic gains over its competitors---achieving a suboptimality that reaches machine precision.
The next best competitor \texttt{Catalyst} only obtains a suboptimality of around $10^{-3}-10^{-4}$.
Moreover, we see \sapph{} exhibits global linear convergence, which is consistent with \cref{thm:conv_str_cvx}.

\subsection{Convergence for non-convex objectives}

In this subsection, we conduct experiments when the objective function $\mathcal{R}(w)$ is non-convex. 
Specifically, we consider two non-convex regularizers and we test on both medium-scale and large-scale data. We evaluate the performance of the algorithm by plotting its normalized training loss versus the effect passes of gradient and computation time. We denote $\mathcal{R}_{\textup{best}}$ as the best training loss across all optimizers and $\mathcal{R}_{\textup{test}}$ as our test loss. 
The following plots show the relative error $(\mathcal{R} - \mathcal{R}_{\textup{best}}) / \mathcal{R}_{\textup{best}}$. 

\subsubsection{Medium-Scale Non-Convex Problems} \label{sub: mnc}

\paragraph{Least-squares regression with a SCAD penalty.}

We consider the problem of least-squares regression with a Smoothly Clipped Absolute Deviation (SCAD) \cite{fan2001variable} penalty. 
SCAD regularization enhances $\ell_1$-based methods by reducing bias for large coefficients while maintaining sparsity and improving model selection consistency \cite{fan2001variable}. 
The SCAD penalty is given by
\begin{align} 
    r(w) = \begin{cases}
        \lambda |w| & |w| \leq \lambda \\
        -\frac{|w|^2 - 2a\lambda |w| + \lambda^2}{2(a-1)} &  \lambda < |w| < a\lambda \\
        \frac{(a+1) \lambda^2}{2} & |w| > a \lambda,
    \end{cases}
\end{align}
where $\lambda > 0, a > 2$ are the regularization parameters controlling sparsity and concavity of penalty. 

\begin{figure}[t]
    \centering
    \includegraphics[scale=0.5]{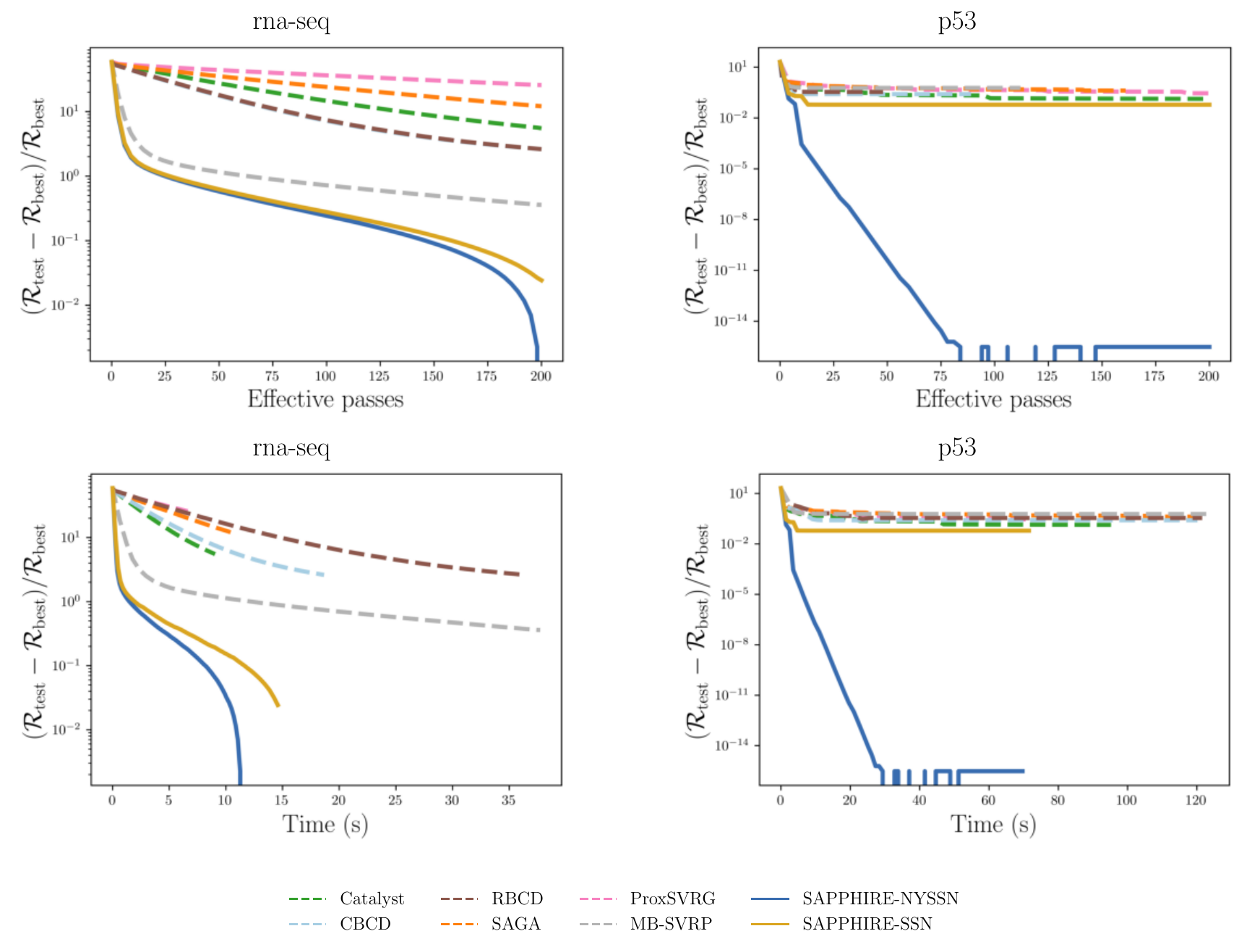}
    \caption{Least-Square regression with SCAD regularization}
    \label{fig:non-convex-1}
\end{figure}

For this experiment, we consider the rnaseq and p53 datasets.
The maximum runtime and number of gradient evaluations is the same as in the medium-scale convex experiments.
The results are presented in \cref{fig:non-convex-1}. \\

\cref{fig:non-convex-1} shows that \sapph{} outperforms all other methods, even though there is no guarantee of an advantage in this setting. 
The most dramatic improvement is observed for the p53 dataset where Sapphire with the NySSN preconditioner converges to point with far better training loss than the competing methods.
Interestingly, the NySSN preconditioner does significantly better on p53 than the SSN preconditioner. 
We attribute this to the fact that in the non-convex setting, the noisier directions of the SSN preconditioner point in bad directions. 
Thus, NySSN's truncation of these directions leads to direct improvements in performance. 

\subsubsection{Large-Scale Non-Convex Problems} \label{sub: lnc}

\begin{figure}[t]
    \centering
    \includegraphics[scale=0.5]{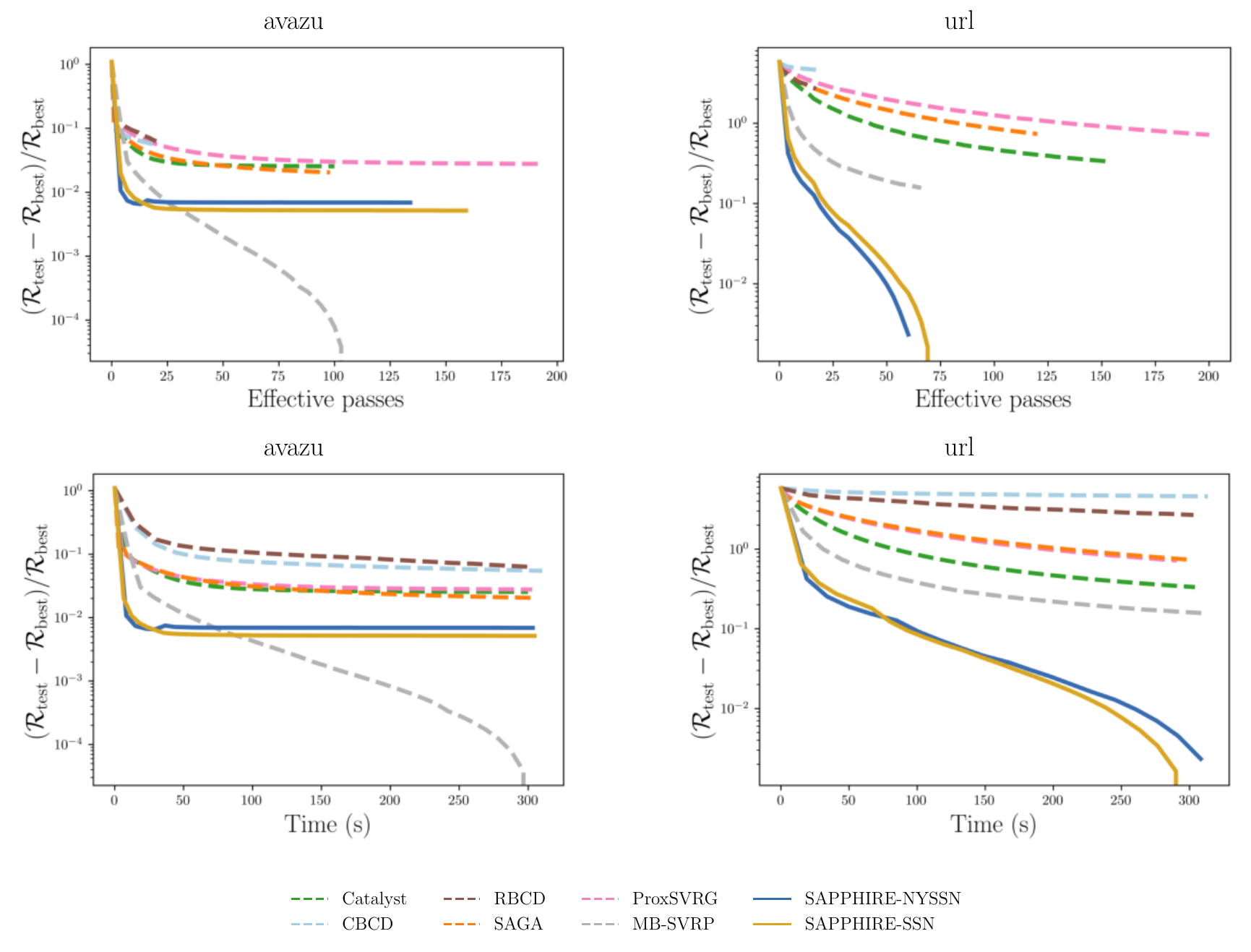}
    \caption{Logistic regression with MCP regularization}
    \label{fig:non-convex-2}
\end{figure}

\paragraph{Logistic regression with a MCP penalty.}
We consider a Minimax Concave Penalty (MCP) \cite{Zhang_2010} regularizer with logistic regression on large-scale datasets avazu and url. This regularization highlights the most relevant features by penalizing large coefficients for less important variables while allowing larger coefficients for key features. We formulate the regularizer as
\begin{align}
    r(w) = \begin{cases}
        \lambda |w| - \frac{w^2}{2 \gamma} & |w| \leq \gamma \lambda \\
        \frac{\gamma \lambda^2}{2} & |w| > \gamma \lambda,
    \end{cases}
\end{align}
where $\lambda > 0, \gamma > 1$ are the regularization parameters controlling the strength and concavity of the penalty. Results appear in \cref{fig:non-convex-2}. \\

\begin{figure}[h]
    \centering
    \includegraphics[scale=0.5]{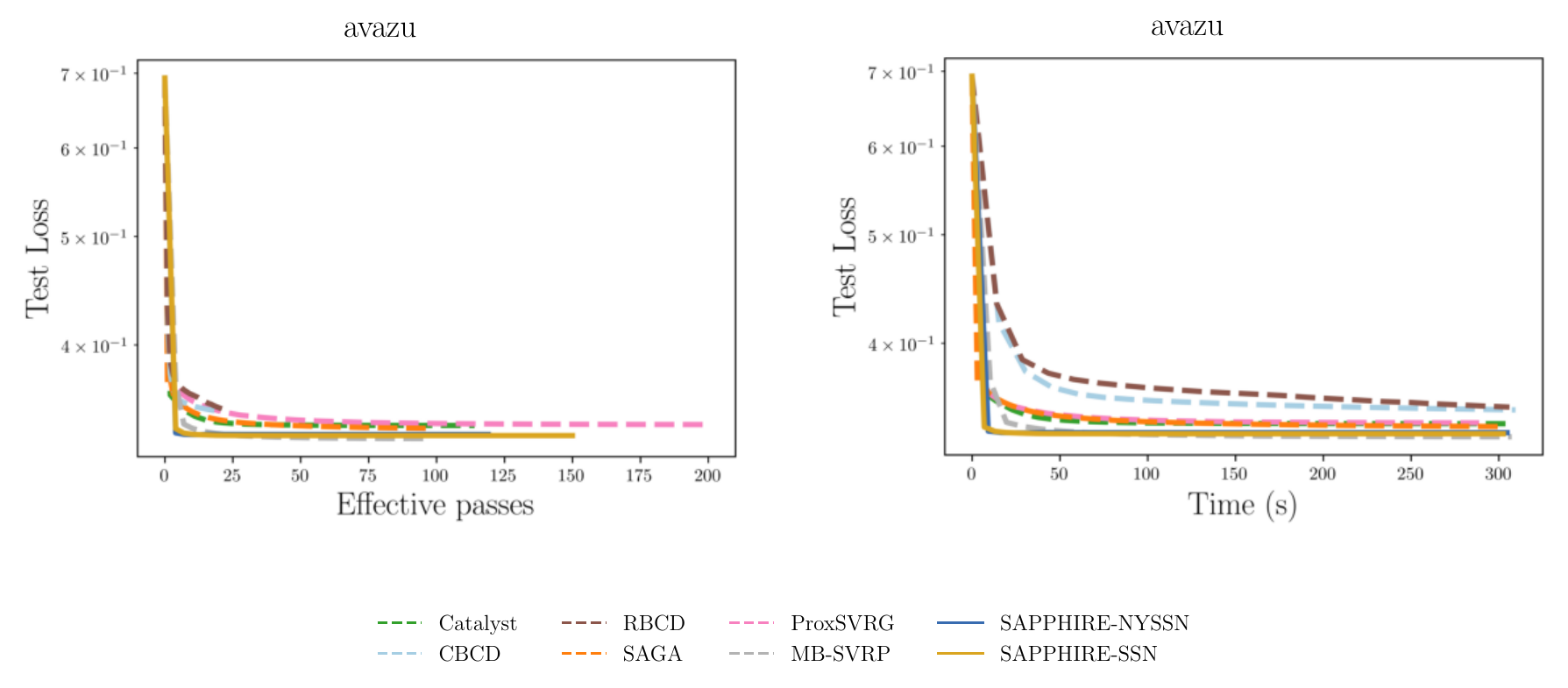}
    \caption{Test Loss for avazu
    \label{fig:test_loss}}
\end{figure}

\sapph{} still shows a solid performance even for ill-conditioned, non-convex, and large-scale problems. 
Both SSN and NySSN show linear convergence on avazu and (after an initial sublinear phase) on url. 
\texttt{MB-SVRP} and \sapph{} converges to different stationary points for avazu. 
Although \texttt{MB-SVRP} has a slightly lower objective value, \cref{fig:test_loss} shows their test loss are not practically significant. 
These experiments demonstrate that \sapph{} often works well even on challenging large-scale non-convex problems. 

\section{Conclusion}
We propose \sapph{}, an optimization algorithm to accelerate large-scale statistical learning for ill-conditioned and non-smooth regularized empirical risk minimization problems. The algorithm leverages sketch-based preconditioning techniques (SSN and NySSN) and the proximal mapping of the regularizer, efficiently approximating curvature information with reduced stochastic variance while maintaining low computational cost. \\

We provide a rigorous theoretical analysis for the convergence of the \sapph{} algorithm, demonstrating global and local linear convergence under quadratic regularity and sublinear convergence under general convex and weak quadratic regular conditions. Empirical results across diverse datasets validate the superior performance of our algorithm in both convergence speed and computational efficiency compared to baseline methods like \texttt{Prox-SVRG} and \texttt{SAGA}. Furthermore, large-scale experiments with convex and non-convex regularizers, including elastic-net and MCP, illustrate the robustness and adaptability of \sapph{} across different problem settings. \\

Therefore, we introduce a robust and efficient framework to address the challenges of ill-conditioned, composite, large-scale optimization problems arising in machine learning. By integrating variance reduction techniques with preconditioned proximal mappings, the \sapph{} algorithm not only improves optimization performance but also offers a scalable and versatile solution for modern data-driven applications. 

\pagebreak
\bibliographystyle{plain}
\bibliography{references.bib}

\appendix
\onecolumn
\renewcommand{\partname}{}
\part{Appendix}

\vspace{10pt}


In this section, we provide the proof for all the lemmas and theorems we present in the main paper, and some auxiliary results which are helpful for the proof. 

\section{Nystr{\"o}m Preconditioner Construction} 

We propose the following algorithm of randomized low-rank approximation to assist the construction of Nystr{\"o}m preconditioner in \cref{pre:nyssn}. 

\begin{algorithm}[h!]
\caption{RandNysApprox
\label{alg:nyssn}}
    \KwIn{Orthogonalized test matrix $\Omega \in \mathbb{R}^{p \times r_H}$, $r_H = \text{rank}(H_{S_H})$, Sketch matrix $M = \widehat \nabla^2 L(w) \Omega \in \mathbb{R}^{p \times r_H}$}

    Compute shift $\nu = \sqrt{p} \cdot \text{eps}(\sigma_{\max}(M))$

    $M_{\nu} = M + \nu \Omega$

    Cholesky decomposition $C = \text{chol}(\Omega^\top M_{\nu})$

    Thin SVD $[\widehat{V}, \Sigma, \sim] = \text{svd}(MC^{-1}, 0)$

    $\widehat{\Lambda} = \max\{0, \Sigma^2 - \nu I\}$
    
    \Return{$\widehat{V}, \widehat{\Lambda}$}
\end{algorithm}

\cref{alg:nyssn} provides the Hessian approximation and construct the Nystr{\"o}m preconditioner in \eqref{eq:nyssn} as $P = \widehat{V} \widehat{\Lambda} \widehat{V}^\top$. Here the function $\text{eps}(\cdot)$ represents the positive distance to the next largest floating point number of the same precision. All eigenvalues of the approximation are non-negative. We apply it in conjunction with a regularizer to ensure positive definiteness. 

\section{Proofs for global convergence of \sapph{}}
In this section, we provide proofs for all results related to the global convergence of \sapph{}.
\subsection{Proof for Lemma \ref{lem:smooth}}
\label{subsec:lem_smooth_pf}
\begin{proof}
By Proposition 3.16 in \cite{promise}, it holds that
\begin{align*}
    \mathbb{E} \| \hat{\nabla} L(w) - \hat{\nabla} L(w^\star) \|_{P^{-1}}^2 & \leq 2 \mathcal L_P \left(L(w) - L(w^\star) - \langle \nabla L(w^\star), w - w^\star\rangle \right).
\end{align*}

Now, by the optimality of $w^\star = \argmin_w \{L(w) + r(w)\}$, there exists $\xi^\star \in \partial r(w^\star)$ such that $\nabla L(w^\star) + \xi^\star = 0$. Thus, by the convexity of $r(w)$, we deduce \begin{align*}
    L(w) - L(w^\star) - \langle \nabla L(w^\star), w - w^\star \rangle &= L(w) - L(w^\star) + \langle \xi^\star, w - w^\star \rangle \\
    & \leq L(w) - L(w^\star) + r(w) - r(w^\star) \\
    &= \mathcal{R}(w) - \mathcal{R}(w^\star). 
\end{align*}

Combining these two results, 
\begin{align*}
    \mathbb{E} \| \hat{\nabla} L(w) - \hat{\nabla} L(w^\star) \|_{P^{-1}}^2 \leq 2 \mathcal L_P [\mathcal{R}(w) - \mathcal{R}(w^\star)]. \\
\end{align*}
\end{proof}

\subsection{Proof for Lemma \ref{lem:var}}
\label{subsec:lem_var_pf}
First, we calculate the expectation of $v_k^{(s)}$ as \begin{align*}
    \mathbb{E} [v_k^{(s)}] &= \mathbb{E} [\hat{\nabla} L(w_k^{(s)})] - \mathbb{E} [\hat{\nabla} L(\hat{w}^{(s)})] + \nabla L(\hat{w}^{(s)}) \\
    &= \nabla L(w_k^{(s)}) - \nabla L(\hat{w}^{(s)}) + \nabla L(\hat{w}^{(s)}) \\
    &= \nabla L(w_k^{(s)}). 
\end{align*}

Building on \cref{lem:smooth}, we derive \begin{align*}
    \mathbb{E} \| v_k^{(s)} - \nabla L(w_k^{(s)}) \|_{(P_k^{(s)})^{-1}}^2 &= \mathbb{E} \| \hat{\nabla} L(w_k^{(s)}) - \hat{\nabla} L(\hat{w}^{(s)}) + \nabla L(\hat{w}^{(s)}) - \nabla L(w_k^{(s)}) \|_{(P_k^{(s)})^{-1}}^2 \\
    & \leq \mathbb{E} \| \hat{\nabla} L(w_k^{(s)}) - \hat{\nabla} L(\hat{w}^{(s)}) \|_{(P_k^{(s)})^{-1}}^2 \\
    & \ \ - \| \nabla L(w_k^{(s)}) - \nabla L(\hat{w}^{(s)}) \|_{(P_k^{(s)})^{-1}}^2 \\
    & \leq \mathbb{E} \| \hat{\nabla} L(w_k^{(s)}) - \hat{\nabla} L(\hat{w}^{(s)}) \|_{(P_k^{(s)})^{-1}}^2 \\
    & \leq 2 \mathbb{E} \| \hat{\nabla} L(w_k^{(s)}) - \hat{\nabla} L(w^\star) \|_{(P_k^{(s)})^{-1}}^2 \\
    & \ \ + 2 \mathbb{E} \| \hat{\nabla} L(\hat{w}^{(s)}) - \hat{\nabla} L(w^\star) \|_{(P_k^{(s)})^{-1}}^2 \\
    & \leq 4 \mathcal L_P [\mathcal{R}(w_k^{(s)}) - \mathcal{R}(w^\star) + \mathcal{R}(\hat{w}^{(s)}) - \mathcal{R}(w^\star)]. \\
\end{align*}
Here, the first inequality uses $\mathbb{E} \| X - \mathbb{E} X \|_A^2 \leq \mathbb{E} \| X \|_A^2$, which is valid for any random variable $X \in \mathbb{R}^d$ and symmetric positive definite matrix $A$. The third inequality uses $\| a + b \|_A^2 \leq 2(\| a \|_A^2 + \| b \|_A^2)$. The last inequality applies \cref{lem:smooth} twice. \\

\subsection{\sapph{}: Global Linear Convergence}
\label{subsection:str_cvx_pf}
We need the following technical result to establish global linear convergence of \sapph{}, which extends Lemma 3 in \cite{xiao2014proximal} to the preconditioned setting.
\begin{lem} \label{lem:add}
    Let $L(w)$ be quadratically regular and $r(w)$ be convex. 
    For any $w \in$ dom(r) and arbitrary $v \in \mathbb{R}^d$, define $\tilde{w} = \prox_{\eta r}^{P} (w - \eta P^{-1}v), g_P = \frac{1}{\eta} P(w - \tilde{w})$, and $\Delta = v - \nabla L(w)$, where $0 < \eta \leq \frac{1}{(1+\zeta) \gamma_{u}}$. 
    Then we have for any $w' \in \mathbb{R}^p$, 
    \begin{align*}
        \mathcal{R}(w') \geq \mathcal{R}(\tilde{w}) + \langle g_P, w'-w \rangle + \frac{\eta}{2} \|g_P\|_{P^{-1}}^2 + \frac{(1-\zeta) \gamma_\ell}{2} \| w'-w \|_{P}^2 + \langle \Delta, \tilde{w}-w'\rangle. 
    \end{align*}
\end{lem}

\begin{proof} We write the proximal update $\tilde{w}$ explicitly as \begin{align*}
    \tilde{w} &= \prox_{\eta r}^{P} (w - \eta P^{-1} v) \\
    &= \argmin_{w'} \left\{\frac{1}{2} \| w' - (w - \eta P^{-1} v) \|_{P}^2 + \eta r(w') \right\}.
\end{align*}
The associated optimality condition states that there exists a $\xi \in \partial r(\tilde{w})$ such that 
\begin{align*}
    P\left(\tilde{w} - (w - \eta P^{-1}v)\right) + \eta \xi = 0.
\end{align*}
and we note that $g_P = P(w - \tilde{w}) / \eta$, so we have $\xi = g_P - v$. 

Applying quadratic regularity of $L$, we can lower bound $L(w)$ by 
\begin{align*}
    L(w) & \geq L(\tilde{w}) - \langle \nabla L(w), \tilde{w} - w \rangle - \frac{(1+\zeta) \gamma_{u}}{2} \| \tilde{w} - w \|_{P}^2 \\ 
    & \geq L(\tilde{w}) - \langle \nabla L(w), \tilde{w} - w \rangle - \frac{1}{2\eta} \| \tilde{w} - w \|_{P}^2.
\end{align*}

By the lower quadratic regularity of $L$ and convexity of $r$, we have for any $w \in$ dom(r) and $w' \in \mathbb{R}^d$, \begin{align*}
    \mathcal{R}(w') &= L(w') + r(w') \\
    & \geq L(w) + \nabla L(w)^\top (w'-w) + \frac{(1-\zeta) \gamma_{\ell}}{2} \| w' - w \|_{P}^2 + R(\tilde{w}) + \xi^\top (w'-\tilde{w}) \\
    & \geq L(\tilde{w}) - \nabla L(w)^\top (\tilde{w} - w) - \frac{1}{2\eta} \| \tilde{w} - w \|_{P}^2 \\
    & \ \ + \nabla L(w)^\top (w'-w) + \frac{(1-\zeta) \gamma_{\ell}}{2} \| w' - w \|_{P}^2 + r(\tilde{w}) + \xi^\top (w'-\tilde{w}) \\
    &= \mathcal{R}(\tilde{w}) + \nabla L(w)^\top (w' - \tilde{w}) + \xi^\top (w'-\tilde{w}) - \frac{1}{2\eta} \| \tilde{w} - w \|_{P}^2 + \frac{(1-\zeta) \gamma_{\ell}}{2} \| w' - w \|_{P}^2.
\end{align*}

Note that $g_P = \frac{1}{\eta} P(w - \tilde{w})$, so we have \begin{align*}
    \frac{1}{2\eta} \| \tilde{w} - w \|_{P}^2 = \frac{1}{2\eta} \cdot \eta^2 \langle P^{-1} g_P, P (P^{-1} g_P)\rangle = \frac{\eta}{2} \langle g_P, P^{-1} g_P \rangle = \frac{\eta}{2} \| g_P \|_{P^{-1}}^2.
\end{align*}

Collect all the inner products on the right-hand-side and denote $\Delta = v - \nabla L(w)$, we have \begin{align*}
     & \langle \nabla L(w), w' - \tilde{w} \rangle + \langle \xi, w'-\tilde{w} \rangle \\
     = & \langle \nabla L(w), w' - \tilde{w} \rangle + \langle g_P - v , w' - \tilde{w}\rangle \\
     = & \langle g_P, w' - \tilde{w} \rangle + \langle v - \nabla L(w), \tilde{w} - w' \rangle \\
     = & \langle g_P, w' - w + w - \tilde{w} \rangle + \langle \Delta, \tilde{w} - w' \rangle \\
     = & \langle g_P, w' - w \rangle + \langle g_P, \eta P^{-1} g_P) \rangle + \langle \Delta, \tilde{w} - w' \rangle \\
     = & \langle g_P, w' - w \rangle + \eta \| g_P \|_{P^{-1}}^2 + \langle \Delta, \tilde{w} - w' \rangle.
\end{align*}

Plugging the derivation of $\frac{1}{2\eta}\| \tilde{w} - w \|_{P}^2$ and $\langle \nabla L(w), w' - \tilde{w} \rangle + \langle \xi, w'-\tilde{w} \rangle$ back for $\mathcal{R}(w')$, we obtain \begin{align*}
    \mathcal{R}(w') & \geq \mathcal{R}(\tilde{w}) + \langle \nabla L(w), w' - \tilde{w} \rangle + \langle \xi, w'-\tilde{w} \rangle - \frac{1}{2\eta} \| \tilde{w} - w \|_{P}^2 + \frac{(1-\zeta) \gamma_{\ell}}{2} \| w' - w \|_{P}^2 \\
    & \geq \mathcal{R}(\tilde{w}) + \langle g_P, w' - w \rangle + \eta \| g_P \|_{P^{-1}}^2 + \langle \Delta, \tilde{w} - w'\rangle - \frac{\eta}{2} \| g_P \|_{P^{-1}}^2 + \frac{(1-\zeta) \gamma_{\ell}}{2} \| w' - w \|_{P}^2 \\
    &= \mathcal{R}(\tilde{w}) + \langle g_P, w'-w \rangle + \frac{\eta}{2} \|g_P\|_{P^{-1}}^2 + \frac{(1-\zeta) \gamma_\ell}{2} \| w'-w \|_{P}^2 + \langle \Delta, \tilde{w}-w' \rangle. 
\end{align*}
\end{proof}

With \cref{lem:add}, we can now begin the proof of \cref{thm:conv_str_cvx}.
The proof is based on a sequence of lemmata.
We begin with the following result, which provides a bound for \sapph{} after one inner iteration.
\begin{lem}[Bound for One Inner Iteration] \label{lem:inner}
   Suppose we are in outer iteration $s$ at inner iteration $k$. 
   Then the following inequality holds  
    \begin{align*}
        \E \left[\| w_{k+1}^{(s)} - w^\star \|_{P_k^{(s)}}^2\right]+ 2\eta \E \left[\mathcal{R}(w_{k+1}^{(s)}) - \mathcal{R}(w^\star) \right] & \leq \| w_{k}^{(s)} - w^\star \|_{P_k^{(s)}}^2 +8 \eta^2 \mathcal L_P [\mathcal{R}(w_k^{(s)}) - \mathcal{R}(w^\star) + \mathcal{R}(\hat{w}^{(s)}) - \mathcal{R}(w^\star)]. 
    \end{align*}
\end{lem}

\begin{proof}
Define the stochastic gradient mapping 
\begin{align*}
    \widehat{G}^{(s)}_k = \frac{1}{\eta}\left(w_k^{(s)} - w_{k+1}^{(s)}\right) = \frac{1}{\eta} \left(w_k^{(s)} - \prox_{\eta r}^P\left(w_k^{(s)} - \eta P_k^{(s)^{-1}} v_k^{(s)}\right)\right),
\end{align*}
so the proximal gradient step can be written as 
\begin{align*}
    w_{k+1}^{(s)} = w_k^{(s)} - \eta \widehat{G}^{(s)}_k. 
\end{align*}
Moreover, we define
\begin{align*}
    \tilde p_k^{(s)} \coloneqq \left(P_k^{(s)}\right)^{-1}v^{(s)}_k, \quad p^{(s)}_k \coloneqq \left(P_k^{(s)}\right)^{-1}\nabla F(w^{(s)}_k).
\end{align*}

Applying the previous relation, we deduce that
\begin{align*}
    \| w_{k+1}^{(s)} - w^\star \|_{P_k^{(s)}}^2 &= \| w_k^{(s)} - \eta \widehat{G}^{(s)}_k - w^\star \|_{P_k^{(s)}}^2 \\
    &= \| w_k^{(s)} - w^\star \|_{P_k^{(s)}}^2 - 2\eta \langle \widehat{G}^{(s)}_k, w_k^{(s)} - w^\star\rangle_{P_k^{(s)}} + \eta^2 \| \widehat{G}^{(s)}_k \|_{P_k^{(s)}}^2.
\end{align*}

Note that our assumptions guarantee $\eta < \frac{1}{4 \mathcal L_P}$. Applying \cref{lem:add} with $w = w_k^{(s)}, v = v_k^{(s)}, \tilde{w} = w_{k+1}^{(s)}, g_P = P_k^{(s)} \widehat{G}^{(s)}_k, w' = w^\star$ and $\Delta^{(s)}_k = v_k^{(s)} - \nabla L(w_k^{(s)})$, we have \begin{align*}
    & - \langle \widehat{G}^{(s)}_k, w_k^{(s)} - w^\star\rangle_{P_k^{(s)}} + \frac{\eta}{2} \| \widehat{G}^{(s)}_k \|_{P_k^{(s)}}^2 \\
    & \leq \mathcal{R}(w^\star) - \mathcal{R}(w_{k+1}^{(s)}) - \frac{(1-\zeta) \gamma_{\ell}}{2} \| w^\star - w_k^{(s)} \|_{P_k^{(s)}}^2 - \langle \Delta^{(s)}_k, w_{k+1}^{(s)} - w^\star \rangle.
\end{align*}

This property of gradient mapping derives the iteration that 
\begin{align*}
    \| w_{k+1}^{(s)} - w^\star \|_{P_k^{(s)}}^2 & \leq \| w_{k}^{(s)} - w^\star \|_{P_k^{(s)}}^2 - \eta (1-\zeta) \gamma_{\ell} \| w_{k}^{(s)} - w^\star \|_{P_k^{(s)}}^2 \\
    & \ \ \ - 2\eta[\mathcal{R}(w_{k+1}^{(s)}) - \mathcal{R}(w^\star)] - 2\eta \langle \Delta^{(s)}_k, w_{k+1}^{(s)} - w^\star\rangle \\
    & \leq \| w_{k}^{(s)} - w^\star \|_{P_k^{(s)}}^2 - 2\eta [\mathcal{R}(w_{k+1}^{(s)}) - \mathcal{R}(w^\star)] - 2\eta \langle \Delta^{(s)}_k, w_{k+1}^{(s)} - w^\star \rangle.
\end{align*}

Next, we bound the quantity $- 2\eta \langle \Delta^{(s)}_k, w_{k+1}^{(s)} - w^\star \rangle$.
Let $\bar{w}_{k+1}^{(s)}$ denote the result of taking a preconditioned proximal gradient step with the full gradient as
\[
    \bar{w}_{k+1}^{(s)} \coloneqq \prox_{\eta r}^P \left(w_{k}^{(s)} - \eta p_k^{(s)}\right).
\] 

Expanding $w_{k+1}^{(s)} - w^\star$ with $\bar{w}_{k+1}^{(s)}$, 
\begin{align*}
    -2\eta \langle \Delta^{(s)}_k, w_{k+1}^{(s)} - w^\star \rangle
    &= -2\eta \langle \Delta^{(s)}_k, w_{k+1}^{(s)} - \bar{w}_{k+1}^{(s)}\rangle  - 2\eta \langle \Delta^{(s)}_k, \bar{w}_{k+1}^{(s)} - w^\star\rangle \\
    & \leq 2\eta \| \Delta^{(s)}_k \|_{P_k^{(s)^{-1}}} 
    \| w_{k+1}^{(s)} - \bar{w}_{k+1}^{(s)} \|_{P_k^{(s)}} - 2\eta \langle \Delta^{(s)}_k, \bar{w}_{k+1}^{(s)} - w^\star\rangle \\ 
    & \leq 2\eta \|\Delta^{(s)}_k \|_{P_k^{(s)^{-1}}} \left\| \left(w_{k}^{(s)} - \eta \tilde{p}_k^{(s)}\right) - \left(w_{k}^{(s)} - \eta p^{(s)}_k\right) \right\|_{P_k^{(s)}} \\
    & \ \ - 2\eta \langle \Delta^{(s)}_k, \bar{w}_{k+1}^{(s)} - w^\star\rangle \\
    &= 2\eta \| \Delta^{(s)}_k \|_{P_k^{(s)^{-1}}} \| \eta P_k^{(s)^{-1}} \Delta^{(s)}_k \|_{P_k^{(s)}} - 2\eta \langle \Delta^{(s)}_k, \bar{w}_{k+1}^{(s)} - w^\star\rangle \\
    &= 2\eta^2 \| \Delta^{(s)}_k \|_{P_k^{(s)^{-1}}}^2 - 2\eta \langle \Delta^{(s)}_k, \bar{w}_{k+1}^{(s)} - w^\star \rangle
\end{align*}
Here, we use Cauchy-Schwarz inequality for the first inequality and non-expansiveness of proximal mapping for the second inequality. 

Combining with the previous result, we have 
\begin{align*}
    \| w_{k+1}^{(s)} - w^\star \|_{P_k^{(s)}}^2 & \leq \| w_{k}^{(s)} - w^\star \|_{P_k^{(s)}}^2 - 2\eta [\mathcal{R}(w_{k+1}^{(s)}) - \mathcal{R}(w^\star)] \\
    & + 2\eta^2 \| \Delta^{(s)}_k \|_{P_k^{(s)^{-1}}}^2 - 2\eta \langle \Delta^{(s)}_k, \bar{w}_{k+1}^{(s)} - w^\star \rangle. 
\end{align*}


Taking the expectation over $v_k^{(s)}$ of both sides of the preceding display and applying \cref{lem:var} obtains
 \begin{align*}
    \E\left[\| w_{k+1}^{(s)} - w^\star \|_{P_k^{(s)}}^2 \right] &= \| w_{k}^{(s)} - w^\star \|_{P_k^{(s)}}^2 - 2\eta \E[\mathcal{R}(w_{k+1}^{(s)}) - \mathcal{R}(w^\star)] \\
    & \ \ + 2\eta^2 \E \left[\| v_k^{(s)} - \nabla L(w_k^{(s)}) \|_{P_k^{(s)^{-1}}}^2 \right] \\
    & \leq \| w_{k}^{(s)} - w^\star \|_{P_k^{(s)}}^2 - 2\eta \mathbb{E}[\mathcal{R}(w_{k+1}^{(s)}) - \mathcal{R}(w^\star) ] \\
    & \ \ + 8 \mathcal L_P \eta^2 [\mathcal{R}(w_k^{(s)}) - \mathcal{R}(w^\star) + \mathcal{R}(\hat{w}^{(s)}) - \mathcal{R}(w^\star)]. \\
\end{align*}
Rearranging the last display, we conclude the desired result.
\end{proof}

\cref{lem:inner} establishes a bound for one inner iteration, which we shall use to establish the following contraction relation for one outer iteration.
\begin{lem}[Bound for One Outer Iteration] \label{lem:outer}
    Suppose we are in outer iteration $s$. 
    Then the output of this outer iteration $\hat w^{(s+1)}$ satisfies
    \begin{align*}
        \mathbb{E}[\mathcal{R}(\hat{w}^{(s+1)})] - \mathcal{R}(w^\star) \leq \left(\frac{1}{(1-\zeta) \gamma_{\ell} \eta (1 - 4\mathcal L_P \eta)m} + \frac{4 \mathcal L_P \eta (m+1)}{(1 - 4\mathcal L_P \eta)m} \right) [\mathcal{R}(\hat{w}^{(s)}) - \mathcal{R}(w^\star)]. 
    \end{align*}
\end{lem}

\textit{Proof.} Applying Lemma \ref{lem:inner} for $k = 0, ..., m-1$, and summing yields
\begin{align*}
    \sum_{k=0}^{m-1}\E[\|w_{k+1}^{(s)}-w^\star\|_{P_k^{(s)}}]+2\eta \sum_{k=0}^{m-1}\E \left[\mathcal{R}(w_{k+1}^{(s)}) - \mathcal{R}(w^\star) \right] &\leq \sum^{m-1}_{k=0}\| w_{k}^{(s)} - w^\star \|^2_{P_k^{(s)}} \\ 
    &+4 \eta \mathcal L_P\sum_{k=0}^{m-1}[\mathcal{R}(w_k^{(s)}) - \mathcal{R}(w^\star) + \mathcal{R}(\hat{w}^{(s)}) - \mathcal{R}(w^\star)]
\end{align*}
Taking the total expectation over the inner iterations and rearranging yields
\begin{align*}
    & \E[\|w_k^{(m)}-w_\star\|^2_{P_k^{(s)}}]+2\eta\E[\mathcal R(w^{(s)}_{k+1})-\mathcal R(w_\star)]+2\eta(1-4\eta \mathcal L_P)\sum_{k=1}^{m-1}\E[\mathcal R(w^{(s)}_{k})-\mathcal R(w_\star)] \\
    &\leq \|\hat w^{(s)}-w_\star\|^2_{P^{(s)}_k}+8(m+1)\eta^2\mathcal L_P(\mathcal R(\hat w^{(s)})-\mathcal R(w^\star)).
\end{align*}
Our choice of $\eta$ implies $2\eta \geq 2\eta(1-4\eta \mathcal L_P)$, yielding
\begin{align*}
    & \E[\|w_k^{(m)}-w_\star\|^2_{P_k^{(s)}}]+2\eta(1-4\eta \mathcal L_P)\sum_{k=1}^{m}\E[\mathcal R(w^{(s)}_{k})-\mathcal R(w_\star)] \leq \|\hat w^{(s)}-w_\star\|^2_{P^{(s)}_k}+8(m+1)\eta^2\mathcal L_P(\mathcal R(\hat w^{(s)})-\mathcal R(w^\star)).
\end{align*}
Rearranging, using the definition of $\hat w^{(s+1)}$ and convexity of $\mathcal R$ yields
\begin{align*}
    \E\left[\mathcal{R}(\hat w^{(s+1)}) - \mathcal{R}(w^\star)\right] & \leq \frac{1}{2\eta m\left(1-4 \eta \mathcal L_P\right)}\| \hat w^{(s)} - w^\star \|_{P_k^{(s)}}^2 + \frac{4 \eta \mathcal L_P(m+1)}{m(1-4\eta \mathcal L_P)}\left(\mathcal{R}(\hat{w}^{(s)}) - \mathcal{R}(w^\star)\right).
\end{align*}

Now, by lower quadratic regularity of $L$ and optimality of $w^\star$, we have
\begin{align*}
    \| \hat{w}^{(s)} - w^\star \|_{P_0^{(s)}}^2 & \leq \frac{2}{(1-\zeta) \gamma_{\ell}}[L(\hat{w}^{(s)}) - L(w^\star)] \\
    & \leq \frac{2}{(1-\zeta) \gamma_{\ell}} [L(\hat{w}^{(s)}) - L(w^\star) + r(\hat{w}^{(s)}) - r(w^\star)] \\
    &= \frac{2}{(1-\zeta) \gamma_{\ell}} [\mathcal{R}(\hat{w}^{(s)}) - \mathcal{R}(w^\star)].
\end{align*}
Here, the second inequality follows from the fact that $r(\hat{w}^{(s)})-r(w^\star)\geq 0$ as $w^\star$ is optimal. 

Combining this with our previous bound, we conclude

\begin{align*}
    \mathbb{E} [\mathcal{R}(\hat{w}^{(s+1)}) - \mathcal{R}(w^\star)] \leq \left(\frac{1}{(1-\zeta) \gamma_{\ell} \eta (1 - 4\eta\mathcal L_P)m} + \frac{4 \eta \mathcal L_P (m+1)}{(1 - 4\eta\mathcal L_P)m} \right) [\mathcal{R}(\hat{w}^{(s)}) - \mathcal{R}(w^\star)]. \\
\end{align*}

The contraction relation in \cref{lem:outer} gives us everything we need to prove \cref{thm:conv_str_cvx}.
\subsection{Proof for Theorem \ref{thm:conv_str_cvx}}
\label{subsec:sapph_conv}
\begin{proof}
Set $\eta = \frac{1}{16 \mathcal L_P}$ and $m = \frac{100 \mathcal L_P}{(1-\zeta) \gamma_{\ell}}$. By \cref{lem:outer}, we perform the recursion and obtain \begin{align*}
    \mathbb{E} \mathcal{R}(\hat{w}^{(s)}) - \mathcal{R}(w^\star) \leq \left(\frac{2}{3}\right)^s (\mathcal{R}(\hat{w}^{(0)}) - \mathcal{R}(w^\star)).
\end{align*}

Therefore, if the number of stages satisfies \begin{align*}
    s \geq 3 \log \left(\frac{\mathcal{R}(\hat{w}^{(0)}) - \mathcal{R}(w^\star)}{\epsilon} \right),
\end{align*}
then we achieve \begin{align*}
    \mathbb{E} \mathcal{R}(\hat{w}^{(s)}) - \mathcal{R}(w^\star) \leq \epsilon. 
\end{align*}

Observing that each stage requires $n + 2m b_g$ component gradient evaluations, we immediately conclude that the total number stochastic gradient evaluations is given by
\begin{align*}
    \bigO \left(\left[n + \frac{\mathcal L_P b_g}{(1-\zeta) \gamma_{\ell}} \right] \log\left(\frac{1}{\epsilon}\right) \right). 
\end{align*}
The rest of the claim follows by substituting in the expression for $\mathcal L_P$ in \cref{lem:smooth}.
\end{proof}

\begin{subsection}{\sapph{}: Sublinear Convergence Analysis}
\label{subsection:conv_cvx_pf}
We now prove \cref{thm:conv_cvx}, which establishes global sublinear convergence of \sapph{} under $\rho$-weak quadratic regularity, which covers the setting when $L(w)$ is only smooth and convex. 
\begin{proof}
Assume we are in outer iteration $s$, then summing the bound in \cref{lem:inner} yields
\begin{align*}
    & \E[\|w_k^{(m)}-w_\star\|^2_{P_k^{(s)}}]+2\eta\E[\mathcal R(w^{(s)}_{k+1})-\mathcal R(w_\star)]+2\eta(1-4\eta \mathcal L_P)\sum_{k=1}^{m-1}\E[\mathcal R(w^{(s)}_{k})-\mathcal R(w_\star)] \\
    &\leq \|\hat w^{(s)}-w_\star\|^2_{P^{(s)}_k}+8(m+1)\eta^2\mathcal L_P(\mathcal R(\hat w^{(s)})-\mathcal R(w^\star)).
\end{align*}
As $\eta = \min\{\frac{1}{4\mathcal L_P(m+2)}, \frac{1}{8(m+2)}\}$ we have that $2\eta \left(1-4 \eta \mathcal L_P\right) \geq \eta^2.$ 
Thus,
\begin{align*}
    & \E[\|\hat w^{(s+1)}-w_\star\|^2_{P_k^{(s)}}]+(2\eta - \eta^2)\E[\mathcal R(\hat w^{(s+1)})-\mathcal R(w_\star)]+\eta^2\sum_{k=1}^{m}\E[\mathcal R(w^{(s)}_{k})-\mathcal R(w_\star)] \\
    &\leq \|\hat w^{(s)}-w_\star\|^2_{P^{(s)}_k}+8(m+1)\eta^2\mathcal L_P(\mathcal R(\hat w^{(s)})-\mathcal R(w^\star)) \\
    &\leq  \|\hat w^{(s)}-w_\star\|^2_{P^{(s)}_k}+(2\eta-\eta^2)(\mathcal R(\hat w^{(s)})-\mathcal R(w^\star)),
\end{align*}
where in the last inequality, we used that value of $\eta$ implies that $2\eta-\eta^2 \geq 8(m+1)\eta^2 \mathcal L_P$.
Thus, the preceding display can be rearranged to yield
\begin{align*}
     & \eta^2\sum_{k=1}^{m}\E[\mathcal R(w^{(s)}_{k})-\mathcal R(w_\star)]\\
    &\leq \|\hat w^{(s)}-w_\star\|^2_{P^{(s)}_k}+(2\eta-\eta^2)(\mathcal R(\hat w^{(s)})-\mathcal R(w^\star)) - \E[\|\hat w^{(s+1)}-w_\star\|^2_{P_k^{(s)}}] - (2\eta - \eta^2)\E[\mathcal R(\hat w^{(s+1)})-\mathcal R(w_\star)].
\end{align*}
Using convexity of $\mathcal R$ this becomes
\begin{align*}
    m\eta^2 \E\left[\mathcal R \left(\frac{1}{m}\sum_{k=1}^{m}w_{k}^{(s)}\right) -\mathcal R(w_\star)\right] &\leq \|\hat w^{(s)}-w_\star\|^2_{P^{(s)}_k}-\E[\|\hat w^{(s+1)}-w_\star\|^2_{P_k^{(s)}}] \\
    &+(2\eta-\eta^2)(\mathcal R(\hat w^{(s)})-\mathcal R(w^\star))-(2\eta - \eta^2)\E[\mathcal R(\hat w^{(s+1)})-\mathcal R(w_\star)].
\end{align*}
Taking the total expectation, summing over all $S$ outer iterations yields, and using convexity of $R$ yields
\begin{align*}
    mS\eta^2 \E\left[\Rc\left(\frac{1}{Sm}\sum_{s=0}^{S-1}\sum_{k=1}^{m}\hat w_k^{(s)}\right) - \Rc(w_\star) \right] \leq \|w_0-w_\star\|_{P_0^{(0)}}^2+(2\eta-\eta^2)\left(\Rc(w_0)-\Rc(w_\star)\right).
\end{align*}
Rearranging, we find that
\begin{align*}
    \E\left[\Rc\left(\frac{1}{Sm}\sum_{s=0}^{S-1}\sum_{k=1}^{m}\hat w_k^{(s)}\right) - \Rc(w_\star) \right] \leq \frac{1}{\eta^2 mS}\|w_0-w_\star\|_{P_0^{(0)}}^2+\frac{1}{mS}\left(\frac{1}{\eta}-1\right)\left(\Rc(w_0)-\Rc(w_\star)\right).
\end{align*}
Using the identity $\frac{1}{\min\left\{a,b\right\}} \leq 1/a+1/b$ for $a,b >0$ yields
\begin{align*}
      \E\left[\Rc\left(\frac{1}{Sm}\sum_{s=0}^{S-1}\sum_{k=1}^{m}\hat w_k^{(s)}\right) - \Rc(w_\star) \right] &\leq \frac{(16\mathcal L_P^2+64)(m+2)^2}{mS}\|w_0-w_\star\|_{P_0^{(0)}}^2+\frac{(4\mathcal L_P +8)(m+2)}{mS}\left(\Rc(w_0)-\Rc(w_\star)\right) \\
      &\leq \frac{3(16\mathcal L_P^2+64)(m+2)}{S}\|w_0-w_\star\|_{P_0^{(0)}}^2+\frac{3(4\mathcal L_P +8)}{S}\left(\Rc(w_0)-\Rc(w_\star)\right).
\end{align*}
Thus, setting $S = \bigO\left(\frac{m \mathcal L_P^2}{ \varepsilon}\right)$ yields
\begin{align*}
    \E\left[\Rc\left(\frac{1}{Sm}\sum_{s=0}^{S-1}\sum_{k=1}^{m}\hat w_k^{(s)}\right) - \Rc(w_\star) \right] &\leq \epsilon\left(\|w_0-w_\star\|_{P_0^{(0)}}^2+\left(\Rc(w_0)-\Rc(w_\star)\right)\right).
\end{align*}
\end{proof}
\end{subsection}

\begin{section}{\sapph{}: Local Convergence Analysis}
\label{sec:sapph_fast_local_convergence}
In this section, we prove \cref{thm:sapph_loc_con}, which shows local condition number-free convergence of \sapph{} in the neighborhood
\[
\Nstar = \left\{w\in \R^p: \|w-w_\star\|_{\nabla^2F(w_\star)}\leq\frac{\varepsilon_0 \nu^{3/2}}{2M}\right\}.
\]
The overall proof strategy is similar to that of other approximate Newton methods. 
Namely, we first show that the iterates remain within $\Nstar$, where the quadratic regularity constants are close to unity. 
Once this has been established, we argue that the output of each stage of \cref{alg:sapphire} contracts to the optimum at a condition number-free rate. 
\subsection{Preliminaries}
We begin by recalling the following technical lemma from \cite{promise}, which shows the following items hold in $\Nstar$: (1) the quadratic regularity constants are close to unity, (2) the Hessians are uniformly close in the Loewner ordering, (3) taking an exact Newton step moves the iterate closer to the optimum in the Hessian norm, (4) $\nabla F_i(w)$, $\nabla F(w)$ are $(1+\varepsilon_0)$ Lipschitz in $\Nstar$.
\begin{lem}
\label{lem:sapph_local_basic_lemmas}
    Let $w,w'\in \Nstar$, and suppose $P$ is a $\varepsilon_0$-spectral approximation constructed at some $w_0\in \Nstar$, 
    then the following items hold.
        \begin{enumerate}
        \item
        \[
        \frac{1}{1+\varepsilon_0}\leq \gamma_{l_\textup{min}}(\Nstar)\leq \gamma_{u_\textup{max}}(\Nstar)\leq (1+\varepsilon_0).
        \]
        \item \[
            (1-\varepsilon_0)\nabla^2 L(w)\preceq \nabla^2 L(w') \preceq (1+\varepsilon_0) \nabla^2L(w). 
            \]
        \item \[
            \|w-w_\star-\nabla^2L(w)^{-1}(\nabla L(w)-\nabla L(w_\star)\|_{\nabla^2 L(w)}\leq \varepsilon_0\|w-w_\star\|_{\nabla^2 L(w)}.
            \]
        \item \begin{align*}
        &\|\nabla L_i(w)-\nabla L_i(w_\star)\|_{\nabla^2 L_i(w')^{-1}}\leq (1+\varepsilon_0)\|w-w_\star\|_{\nabla^2 L_i(w')}, \quad \text{for all $i\in [n]$},\\
        & \|\nabla L(w)-\nabla L(w_\star)\|_{\nabla^2 L(w')^{-1}}\leq (1+\varepsilon_0)\|w-w_\star\|_{\nabla^2 F(w')}.
        \end{align*}
\end{enumerate}
\end{lem}

\subsection{Controlling the error in the stochastic gradient}
Similar to the global convergence analysis, it is essential that the deviation of the variance-reduced gradient from the exact gradient goes to zero as we approach $w_\star$.
Thus, our analysis begins with the following lemma, which gives a high probability bound for the preconditioned gradient error. 
It provides a local analog of \cref{lem:var}.  
\begin{lem}
\label{lem:grad_err_bnd}
    Let $\beta_g\in (0,1)$. If $w_k^{(s)} \in \Nstar$  and $v^{(s)}_k$ is constructed with batchsize $b_g = \bigO\left(\frac{\tau_{\star}^{\nu}(\Nstar)\log(\frac{1}{\delta})}{\beta^2_g}\right)$, then with probability at least $1-\delta$
    \[\|v_k^{(s)}-\nabla L(w_k^{(s)})\|_{P^{-1}}\leq \beta_g\left(\|w_k^{(s)}-w_\star\|_{P}+\|\hat w^{(s)}-w_\star\|_{P}\right).\]
\end{lem}

\begin{proof}
    Let $X_i = \nabla^2 L(w_\star)^{-1/2}\left(\nabla L_i(w^{(s)}_k)-\nabla L_i(\hat w^{(s)})- \left(\nabla L(w_k^{(s)})-\nabla L(\hat w^{(s)})\right)\right)$. 
    By definition of $X_i$, 
    \begin{align*}
        \nabla^2 L(w_\star)^{-1/2}\left(v_k^{(s)}-\nabla L(w^{(s)}_k)\right) = \frac{1}{b_g}\sum_{i \in \mathcal B}X_i \coloneqq X.
    \end{align*}
    Observe that $\|X\| = \|v_k^{(s)}-\nabla L(w_k^{(s)})\|_{\nabla^2L(w_\star)^{-1}}$, and $\E[X] = 0$ by definition of the variance-reduced gradient.
    Therefore, we can control $\|v^{(s)}_k-\nabla L(w_k^{(s)})\|_{\nabla^2 L(w_\star)^{-1}}$ by a concentration argument similar to \cite{promise}.
    We can then convert the result to the $(P^{-1},P)$-dual norm pair by applying \cref{lem:sapph_local_basic_lemmas}. \\

    We shall use Bernstein's inequality for vectors to bound $\|X\|$ with high probability.
    In order to apply this variant of Bernstein's inequality, we must establish bounds on $\|X_i\|$ and $\E\|X_i\|^2.$\\
    We begin by bounding $\|X_i\|$. 
    To this end, observe that,
    \begin{align*}
        \|X_i\|^2 & \overset{(1)}{\leq} 2\|\nabla L_i(w^{(s)}_k)-\nabla L_i(\hat w^{(s)})\|^2_{\nabla^2 L(w_\star)^{-1}}+2\|\nabla L(w^{(s)}_k)-\nabla L(\hat w^{(s)})\|^2_{\nabla^2 L(w_\star)^{-1}} \\
        & \overset{(2)}{\leq} 4\tau_\star(\Nstar)^2(1+\varepsilon_0)^2\|w_k^{(s)}-\hat w^{(s)}\|^2_{\nabla^2 L(w_\star)} \\
        &\leq 8\tau_\star(\Nstar)^2(1+\varepsilon_0)^2\left(\|w_k^{(s)}-w_\star\|^2_{\nabla^2 L(w_\star)}+\|\hat w^{(s)}-w_\star\|^2_{\nabla^2 L(w_\star)}\right).
    \end{align*}
    Here $(1)$ uses $\|x+y\|^2\leq 2\|x\|^2+2\|y\|^2$, and $(2)$ uses \cref{lem:rho_dissim} and item 4 of \cref{lem:sapph_local_basic_lemmas}.
    Taking the square root on both sides yields 
    \[
    \|X_i\| \leq 2\sqrt{2}\tau_\star(\Nstar)(1+\varepsilon_0)\left(\|w_k^{(s)}-w_\star\|_{\nabla^2 L(w_\star)}+\|\hat w^{(s)}-w_\star\|_{\nabla^2 L(w_\star)}\right).
    \]
    This establishes the required bound on $\|X_i\|$.
    We now turn to bounding $\E \|X_i\|^2$.
    To begin, observe that an argument similar to the one in \cref{lem:var} yields
    \[
      \mathbb E\|X_i\|^2 \leq 2\mathbb E\|\nabla L_i(w^{(s)}_k)-\nabla L_i(w_\star)\|_{\nabla^2L(w_\star)^{-1}}+2\mathbb E\|\nabla L_i(\hat w^{(s)})-\nabla L_i(w_\star)\|_{\nabla^2L(w_\star)^{-1}}.
    \]
    Again using \cref{lem:rho_dissim} and \cref{lem:sapph_local_basic_lemmas}, we obtain
    \begin{align*}
        & 2\mathbb E\|\nabla L_i(w^{(s)}_k)-\nabla L_i(w_\star)\|_{\nabla^2L(w_\star)^{-1}}+2\mathbb E\|\nabla L_i(\hat w^{(s)})-\nabla L_i(w_\star)\|_{\nabla^2L(w_\star)^{-1}} \\ 
        &\leq 2\tau_\star(\Nstar)\mathbb E\|\nabla L_i(w^{(s)}_k)-\nabla L_i(w_\star)\|_{\nabla^2L_i(w_\star)^{-1}}+2\tau_\star(\Nstar)\mathbb E\|\nabla L_i(\hat w^{(s)})-\nabla L_i(w_\star)\|_{\nabla^2L_i(w_\star)^{-1}}\\
        &\leq 2\tau_\star(\Nstar)(1+\varepsilon_0)\mathbb E\left(L_i(w_k^{(s)})-L_i(w_\star)-\langle \nabla L_i(w_\star), w_k^{(s)}-w_\star\rangle\right)\\
        &+2\tau_\star(\Nstar)(1+\varepsilon_0)\mathbb E\left(L_i(\hat w^{(s)})-L_i(w_\star)-\langle \nabla L_i(w_\star), \hat w^{(s)}-w_\star\rangle\right)\\
        &= 2\tau_\star(\Nstar)(1+\varepsilon_0)\left(L(w_k^{(s)})-L(w_\star)-\langle \nabla L(w_\star), w_k^{(s)}-w_\star\rangle\right)\\
        &+2\tau_\star(\Nstar)(1+\varepsilon_0)\left(L(\hat w^{(s)})-L(w_\star)-\langle \nabla L(w_\star), \hat w^{(s)}-w_\star\rangle\right)\\
        &\leq 2\tau_\star(\Nstar)(1+\varepsilon_0)^2\left(\|w^{(s)}_k-w_\star\|_{\nabla^2 L(w_\star)}+\|\hat w^{(s)}-w_\star\|_{\nabla^2 L(w_\star)}\right).     
   \end{align*}
   Hence, the scaled gradient residual $X_i$ satisfies
   \[
   \mathbb E\|X_i\|^2 \leq 2\tau_\star(\Nstar)(1+\varepsilon_0)^2\left(\|w^{(s)}_k-w_\star\|_{\nabla^2 L(w_\star)}+\|\hat w^{(s)}-w_\star\|_{\nabla^2 L(w_\star)}\right).
   \]
   
   Having bounded $\|X_i\|$ and $\E\|X_i\|^2$, we can apply Lemma 27 from \cite{promise} with $b_g = \bigO\left(\frac{\tau_\star(\Nstar)\log\left(\frac{1}{\delta}\right)}{\beta_g^2}\right)$ to reach 
   \[
   \|v_k^{(s)}-\nabla L(w_k^{(s)})\|_{\nabla^2 F(w_\star)^{-1}} \leq \frac{\beta_g}{4}\left(\|w_k^{(s)}-w_\star\|_{\nabla^2 F(w_\star)}+\|\hat w^{(s)}-w_\star\|_{\nabla^2 F(w_\star)}\right). 
   \]
   Converting to preconditioned norms via \cref{lem:sapph_local_basic_lemmas}, this becomes
   \[
      \|v_k^{(s)}-\nabla L(w_k^{(s)})\|_{P^{-1}}\leq \beta_g\left(\|w_k^{(s)}-w_\star\|_{P}+\|\hat w^{(s)}-w_\star\|_{P}\right).
   \]
\end{proof}

\subsection{Establishing a one iteration contraction}
With \cref{lem:grad_err_bnd} in hand, we now establish a contraction relation for iterates in any outer iteration $s$.
This lemma guarantees the \sapph{} iterates remain in $\Nstar$, essential for showing condition number-free local convergence.
\begin{lem}\label{lem:next_iter_bound}
    Let $w_k^{(s)}\in \Nstar$, and $\beta_g\in (0,1)$. Suppose the gradient batchsize satisfies $b_g = \bigO\left(\frac{\tau_\star^{\nu}(\Nstar)\log\left(\frac{k+1}{\delta}\right)}{\beta^2_g}\right)$. 
    Then with probability at least $1-\frac{\delta}{(k+1)^2}$
    \begin{enumerate}
        \item $\|\Delta_{k+1}^{(s)}\|_{\nabla^2F(w_\star)}\leq \frac{3}{4}\|\Delta^{(s)}_k\|_{\nabla^2F(w_\star)}+\frac{7}{48} \|\Delta^{(s)}_0\|_{\nabla^2F(w_\star)}$
        \item $w^{(s)}_{k+1} \in \Nstar$.
    \end{enumerate}
\end{lem}
\begin{proof}
Let $\Delta_{k+1}^{(s)} = \prox_r^{P}\left(w^{(s)}_k-P^{-1}\nabla L(w^{(s)}_k)\right)-w_\star$. 
We begin with the following inequality,
\begin{align*}
    \left\|\Delta_{k+1}^{(s)}\right\|_P &= \left\|\prox_r^{P}\left(w_k-P^{-1}v_k^{(s)}\right)-w_\star\right\|_P = \left\|\prox_r^{P}\left(w_k-P^{-1}v^{(s)}_k\right)-\prox_r^{P}\left(w_\star-P^{-1}\nabla L(w_\star)\right)\right\|_P \\
    &\leq \left\|\left(w_k-P^{-1}v_k^{(s)}\right)-\left(w_\star-P^{-1}\nabla L(w_\star)\right)\right\|_P \\
    &= \left\|P(w_k-w_\star)-(\nabla L(w_k)-\nabla L(w_\star))+ \nabla L(w_k)-v_k^{(s)}\right\|_{P^{-1}} \\
    &\leq \left\|P(w^{(s)}_k-w_\star)-(\nabla L(w^{(s)}_k)-\nabla L(w_\star))\right\|_{P^{-1}}+\left\|v^{(s)}_k-\nabla L(w^{(s)}_k)\right\|_{P^{-1}},
\end{align*}
where in the second inequality, we used the non-expansiveness of the scaled proximal mapping.
The preceding display consists of two terms. 
The first term represents the error in the approximate Taylor expansion
\[
\nabla L(w^{(s)}_k)-\nabla L(w_\star) \approx P(w^{(s)}_k-w_\star).
\]
The second term measures the deviation of the stochastic gradient from the exact gradient. 
Using \cref{lem:grad_err_bnd}, the second term can be bounded as,
\[
\beta_g\left(\|\Delta_k^{(s)}\|_P+\|\Delta_0^{(s)}\|_P\right).
\]
Thus, we now turn to bounding the Taylor error term.  
To this end, observe that the triangle inequality yields
\begin{align*}
    \left\|P(w^{(s)}_k-w_\star)-(\nabla L(w^{(s)}_k)-\nabla L(w_\star))\right\|_{P^{-1}} &\leq
  \left\|\nabla^2 L(w_k^{(s)})(w^{(s)}_k-w_\star)-(\nabla L(w^{(s)}_k)-\nabla L(w_\star))\right\|_{P^{-1}}\\ 
  &+\left\|\left(P-\nabla^2 L(w_k^{(s)})\right)(w_k^{(s)}-w_\star)\right\|_{P^{-1}}.
\end{align*}
The first term in this inequality is the exact Taylor expansion error, while the second term represents the error in approximating the Hessian.
We can bound the first term using \cref{lem:sapph_local_basic_lemmas} as follows,
\begin{align*}
   \left\|\nabla^2 L(w_k^{(s)})(w^{(s)}_k-w_\star)-(\nabla L(w^{(s)}_k)-\nabla L(w_\star))\right\|_{P^{-1}} & \overset{(1)}{\leq} \frac{1}{\sqrt{1-\varepsilon_0}}  \left\|\nabla^2 L(w_k^{(s)})(w^{(s)}_k-w_\star)-(\nabla L(w^{(s)}_k)-\nabla L(w_\star))\right\|_{\nabla^2 L(w^{(s)}_k)^{-1}}  \\
   &= \frac{1}{\sqrt{1-\varepsilon_0}}\|w^{(s)}_k-w_\star - \nabla^2 L(w_k^{(s)})^{-1}(\nabla L(w_k^{(s)})-\nabla L(w_\star))\|_{\nabla^2 L(w^{(s)}_k)} \\
   &\overset{(2)}{\leq} \frac{\varepsilon_0}{\sqrt{1-\varepsilon_0}}\|\Delta_k^{(s)}\|_{\nabla^2 L(w^{(s)}_k)} \\
   &\overset{(3)}{\leq} \varepsilon_0\sqrt{\frac{1+\varepsilon_0}{1-\varepsilon_0}}\|\Delta_k^{(s)}\|_{P} \\
   & \overset{(4)}{\leq} 2\varepsilon_0 \|\Delta_k^{(s)}\|_{P}.
\end{align*}
Here (1) uses item 1 of \cref{lem:sapph_local_basic_lemmas}, (2) uses item 2 of \cref{lem:sapph_local_basic_lemmas}, (3) uses item of \cref{lem:sapph_local_basic_lemmas} again, and (4) uses $\varepsilon_0 \leq \frac{1}{6}$.

We can also bound Hessian approximation error term via \cref{lem:sapph_local_basic_lemmas}.
Indeed, 
\begin{align*}
    \left\|\left(P-\nabla^2 L(w_k^{(s)})\right)(w_k^{(s)}-w_\star)\right\|_{P^{-1}} & =\left\|P^{1/2}\left(I-P^{-1/2}\nabla^2 F(w_k^{(s)})P^{-1/2}\right)P^{1/2}(w_k^{(s)}-w_\star)\right\|_{P^{-1}} \\
    &= \left\|\left(I-P^{-1/2}\nabla^2 F(w_k^{(s)})P^{-1/2}\right)P^{1/2}(w_k^{(s)}-w_\star)\right\| \\
    &\leq  \left\|I-P^{-1/2}\nabla^2 F(w_k^{(s)})P^{-1/2}\right\| \left\|w_k^{(s)}-w_\star\right\|_P \\
    &\leq \varepsilon_0 \|w_k^{(s)}-w_\star\|_P = \varepsilon_0 \|\Delta_k^{(s)}\|_P,
\end{align*}
where the last inequality uses item 2 of \cref{lem:sapph_local_basic_lemmas}.
Putting together the two bounds, we find the approximate Taylor error term satisfies
\[
 \left\|P(w^{(s)}_k-w_\star)-(\nabla L(w^{(s)}_k)-\nabla L(w_\star))\right\|_{P^{-1}} \leq 3\varepsilon_0\|\Delta_k^{(s)}\|_P.
\]

Combining the bounds on the approximate Taylor error and the error in the stochastic gradient, we deduce 
\[
    \left\|\Delta_{k+1}^{(s)}\right\|_P \leq (\beta_g+3\varepsilon_0)\| \Delta_k^{(s)}\|_P+\beta_g\|\Delta_0^{(s)}\|_P.
\]
Now, converting norms yields
\begin{align*}
    \left\|\Delta_{k+1}^{(s)}\right\|_{\nabla^2 L(w_\star)} &\leq (1+\varepsilon_0)(\beta_g+3\varepsilon_0)\| \Delta_k^{(s)}\|_{\nabla^2 L(w_\star)}+\beta_g(1+\varepsilon_0)\|\Delta_0^{(s)}\|_{\nabla^2 L(w_\star)} \\
    &\leq \frac{3}{4}\| \Delta_k^{(s)}\|_{\nabla^2 L(w_\star)}+\frac{7}{48}\|\Delta_0^{(s)}\|_{\nabla^2 L(w_\star)}.
\end{align*}
\end{proof}

\subsection{Showing convergence for one stage}
Now that we have established the iterates produced by \sapph{} remain in $\Nstar$, we can establish the convergence rate for one stage.
\begin{lem}[One-stage analysis]
\label[lemma]{lem:sksvrg_local_one_stage_bnd}
    Let $\hat w^{(s)}\in \Nstar$. Run \cref{alg:sapphire} with $m = 10$ inner iterations and gradient batchsize satisfies $b_{g} =\bigO\left(\tau_\star^{\nu}(\Nstar)\log\left(\frac{m+1}{\delta}\right)\right)$. 
    Then with probability at least $1-\delta$,
    \begin{enumerate}
        \item $\hat w^{(s+1)} \in \mathcal N_{\frac{2}{3}\varepsilon_0}(w_\star)$.
        \item  $\|\hat w^{(s+1)}-w_\star\|_{\nabla^2 L(w_\star)}\leq \frac{2}{3}\|\hat w^{(s)}-w_\star\|_{\nabla^2 L(w_\star)}$.
    \end{enumerate}
\end{lem}
\begin{proof}
    As $\hat{w}^{(s)}\in \Nstar$, it follows by union bound that the conclusions of \cref{lem:next_iter_bound} hold for all $w_{k}^{(s)}$, where  $k\in \{0,\dots, m-1\}$, with probability at least 
    \[
    1-\sum^{m-1}_{k=0}\frac{\delta}{(m+1)^2} = 1-\frac{m}{(m+1)^2}\delta\geq 1-\delta. 
    \]
    Consequently, applying \cref{lem:next_iter_bound}, 
    \[
    \|\Delta_{m}^{(s)}\|_{\nabla^2L(w_\star)}\leq \frac{3}{4}\|\Delta^{(s)}_{m-1}\|_{\nabla^2L(w_\star)}+\frac{7}{48}\|\Delta^{(s)}_{0}\|_{\nabla^2L(w_\star)}.
    \] 
    Now recursively applying the relation in the previous display, and using $m = 10 > \frac{\log(1/15)}{\log(3/4)}$, we reach 
    \begin{align*}
        &\|\Delta_{m}^{(s)}\|_{\nabla^2L(w_\star)} \leq \left(\frac{3}{4}\right)^{m}\|\Delta^{(s)}_0\|_{\nabla^2L(w_\star)}+\left(\sum_{k=0}^{m-1}\left(\frac{3}{4}\right)^{k}\right)\frac{7}{48}\|\Delta^{(s)}_0\|_{\nabla^2F(w_\star)}\\
        &\leq \frac{1}{15}\|\Delta^{(s)}_0\|_{\nabla^2L(w_\star)}+\frac{7}{48(1-\frac{3}{4})}\|\Delta^{(s)}_0\|_{\nabla^2L(w_\star)}\\ 
        &= \left(\frac{1}{15}+\frac{7}{12}\right)\|\Delta^{(s)}_0\|_{\nabla^2L(w_\star)} \leq 
       \frac{2}{3}\|\Delta^{(s)}_0\|_{\nabla^2L(w_\star)}.
    \end{align*}
    Hence $\hat w^{(s+1)} = w_m^{(s)}\in \mathcal N_{\frac{2}{3}\varepsilon_0}(w_\star)$. 
\end{proof}

We now have everything we need to prove \cref{thm:sapph_loc_con}.
\subsection{Proof for Theorem \ref{thm:sapph_loc_con}}
\label{subsec:sapph_loc_conv}

By \cref{lem:sksvrg_local_one_stage_bnd}, we perform the recursion and obtain
\begin{align*}
    \|\hat w^{(s)}-w_\star\|_{\nabla^2 L(w_\star)} \leq \left(\frac{2}{3}\right)^s \|\hat w^{(0)}-w_\star\|_{\nabla^2 L(w_\star)}.
\end{align*}

Therefore, with $\varepsilon_0 \in (0, 1/6]$, if the number of stages satisfies \begin{align*}
    s \geq 2\log \left(\frac{1}{\epsilon} \right) \geq 3 \log \left(\frac{\|\hat w^{(0)}-w_\star\|_{\nabla^2 L(w_\star)}}{\epsilon} \right),
\end{align*}
then we achieve \begin{align*}
    \|\hat w^{(s)}-w_\star\|_{\nabla^2 L(w_\star)}\leq \epsilon.
\end{align*}

Observing that each stage requires $n + 2m b_g$ component gradient evaluations, and that $\tau^{\rho}(\Nstar) \leq n$ (recall \cref{lem:rho_dissim}), we immediately conclude that the total number stochastic gradient evaluations is given by
\begin{align*}
    \bigO \left(\left[n + \bigOt\left(\tau^{\rho}(\Nstar)\log\left(\frac{1}{\delta}\right)\right) \right] \log\left(\frac{1}{\epsilon}\right) \right) = \bigO \left(n \log \left(\frac{1}{\epsilon} \right) \right). 
\end{align*}

This completes the proof. 

\end{section}


\end{document}